\pdfoutput=1

\documentclass[11pt]{article}

\usepackage[]{emnlp2023}

\usepackage{times}
\usepackage{latexsym}

\usepackage[T1]{fontenc}

\usepackage[utf8]{inputenc}

\usepackage{microtype}

\usepackage{color,soul,hyperref} 
\usepackage{enumitem} 
\usepackage{amsmath,amssymb}
\usepackage{graphicx} 
\usepackage{svg}
\usepackage{booktabs}
\usepackage{multirow}

\newcommand{\qed}{\hfill\square}

\title{A State-Vector Framework for Dataset Effects}

\author{Esmat Sahak$^{1,2}$, Zining Zhu$^{1,2}$, Frank Rudzicz$^{1,2,3}$ \\ $^1$ University of Toronto, $^2$ Vector Institute for Artificial Intelligence, $^3$ Dalhousie University \\ \texttt{esmat.sahak@mail.utoronto.ca}, \texttt{zining@cs.toronto.edu}, \texttt{frank@dal.ca} }

\begin{document}
\maketitle
\begin{abstract}
The impressive success of recent deep neural network (DNN)-based systems is significantly influenced by the high-quality datasets used in training. However, the effects of the datasets, especially how they interact with each other, remain underexplored. We propose a state-vector framework to enable rigorous studies in this direction. This framework uses idealized probing test results as the bases of a vector space. This framework allows us to quantify the effects of both standalone and interacting datasets. We show that the significant effects of some commonly-used language understanding datasets are characteristic and are concentrated on a few linguistic dimensions. Additionally, we observe some ``spill-over'' effects: the datasets could impact the models along dimensions that may seem unrelated to the intended tasks. Our state-vector framework paves the way for a systematic understanding of the dataset effects, a crucial component in responsible and robust model development.
\end{abstract}

\section{Introduction}
In recent years, data-driven systems have shown impressive performance on a wide variety of tasks and massive, high-quality data is a crucial component for their success \citep{brown_language_2020,hoffmann2022training}. Currently, the availability of language data grows much more slowly than the computation power (approximately at Moore's law), raising the concern of ``data exhaustion'' in the near future \citep{villalobos_will_2022}. This impending limitation calls for more attention to study the quality and the effects of data.

The data-driven systems gain linguistic abilities on multiple levels ranging from syntax, semantics, and even some discourse-related abilities during the training procedures \citep{liu_probing_2021}. The training procedures almost always include multiple datasets -- usually there is a ``pre-training'' phase and a ``fine-tuning'' phase, where the model developers apply different datasets. Model developers use large corpora that may include multiple existing datasets \citep{sun_ernie_2021,wei2021finetuned,brown_language_2020}. How these data relate to the progress in linguistic ability is not systematically studied yet. Each dataset has desired effects, but does it have some under-specified effects? When using multiple datasets together, do they have undesired, interaction effects? These questions become more contingent as the engineering of datasets becomes more essential, yet no existing framework allows convenient quantification of these dataset effects.

\begin{figure}[t]
    \centering
    \includegraphics[width=\linewidth]{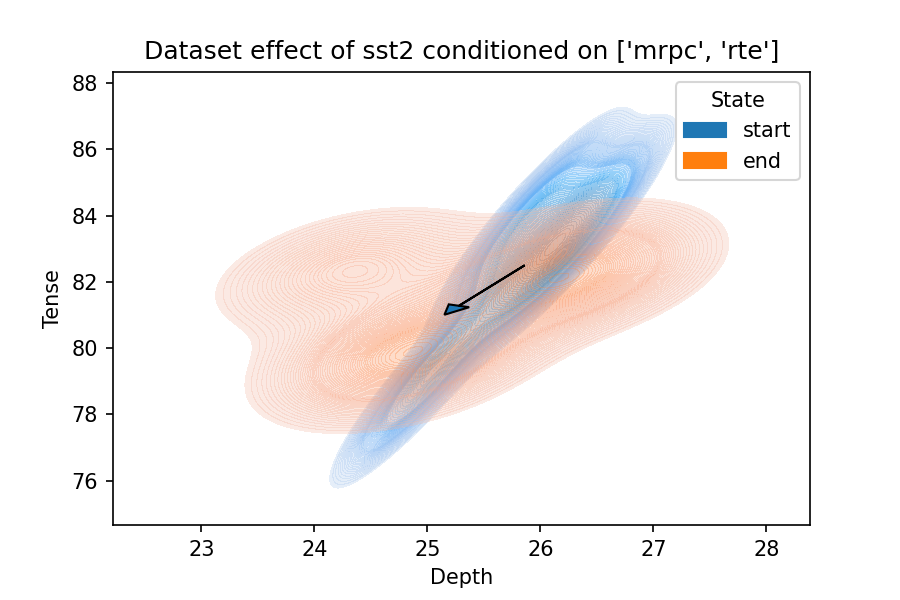}
    \caption{An example of the \textit{individual} dataset effect. This figure shows how two linguistic abilities of RoBERTa, as characterized by the accuracies of two probing tasks: syntactic tree depth and tense, are changed by fine-tuning the SST2 dataset.}
    \label{fig:individual_effect_example}
\end{figure}

Probing provides convenient frameworks to study the linguistic abilities of DNN systems from multiple perspectives. Probing analyses show how the components of DNNs demonstrate linguistic abilities  \citep{tenney_BERT_2019,rogers_primer_2020,belinkov_probing_2021,niu-et-al-2022-BERT}. The probing results are relevant to the model's ability to the extent that the probing accuracies can predict the model's downstream performance \citep{zhu_predicting_2022}. These findings make probing classification a promising candidate for setting up a framework to describe the effects of datasets.

In this paper, we set up a state-vector framework for describing the dataset effects. We formalize idealized probes, which give the true linguistic ability of the DNN model in a state of training. Then, we derive two terms, individual and interaction effects, that describe the dataset effects along multiple dimensions of linguistic abilities. A benefit of our state-vector framework is that it allows a convenient setup of statistical tests, supporting a rigorous interpretation of how the datasets affect the models.

The state framework allows us to frame transitions and set up experiments efficiently. Many frequently-used datasets have ``spill-over'' effects besides the purposes they are originally collected to achieve. Additionally, the interaction effects are concentrated and characteristic. Our framework provides a systematic approach to studying these effects of the datasets, shedding light on an aspect of model developments that deserves more attentions. All scripts, data, and analyses are available at our \href{https://github.com/esmatsahak/EMNLP-2023_A-State-Vector-Framework-for-Dataset-Effects_Repository}{GitHub repository}.


\section{Related Works}
\paragraph{Understanding the datasets}
Recent work has investigated various properties of datasets. \citet{swayamdipta-etal-2020-dataset} used the signals of the training dynamics to map individual data samples onto ``easy to learn'', ``hard to learn'', and ``ambiguous'' regions. \citet{ethayarajh_2022_dataset} used an aggregate score, predictive $\mathcal{V}$-information \citep{xu_theory_2020}, to describe the difficulty of datasets. 
Some datasets can train models with complex decision boundaries. From this perspective, the complexity can be described by the extent to which the data samples are clustered by the labels, which can be quantified by the Calinski-Habasz index. Recently, \citet{jeon2022sanity} generalized this score to multiple datasets.
We also consider datasets in an aggregate manner but with a unique perspective. With the help of probing analyses, we are able to evaluate the effects of the datasets along multiple dimensions, rather than as a single difficulty score. 


\paragraph{Multitask fine-tuning}
\citet{mosbach-etal-2020-interplay-fine} studied how fine-tuning affects the linguistic knowledge encoded in the representations. Our results echo their finding that fine-tuning can either enhance or remove some knowledge, as measured by probing accuracies. Our proposed dataset effect framework formalizes the study in this direction.
\citet{aroca-ouellette_losses_2020} studied the effects of different losses on the DNN models and used downstream performance to evaluate the multidimensional effects. We operate from a dataset perspective and use the probing performance to evaluate the multidimensional effects.
\citet{weller-etal-2022-use} compared two multitask fine-tuning settings, sequential and joint training. The former can reach higher transfer performance when the target dataset is larger in size. We use the same dataset size on all tasks, so either multitask setting is acceptable. 

\section{A State-Vector framework}
This section describes the procedure for formulating a state-vector framework for analyzing dataset effects in multitask fine-tuning.

\paragraph{An abstract view of probes} There have been many papers on the theme of probing deep neural networks. The term ``probing'' contains multiple senses. A narrow sense refers specifically to applying {\em post-hoc} predictions to the intermediate representations of DNNs. A broader sense refers to examination methods that aim at understanding the intrinsics of the DNNs. We consider an abstract view, treating a probing test as a map from multi-dimensional representations to a scalar-valued test result describing a designated aspect of the status of the DNN. Ideally, the test result faithfully and reliably reflects the designated aspect. We refer to the probes as \textit{idealized} probes henceforth.

\paragraph{Idealized probes vectorize model states} Training DNN models is a complex process consisting of a sequence of states. In each state $S$, we can apply a battery of $K$ \textit{idealized} probes to the DNN and obtain a collection of results describing the linguistic capabilities of the model's state at timestep $\mathcal{T}=[T_1, T_2, ... T_K]$. In this way, the state of a DNN model during multitask training can be described by a \textit{state vector} $\mathbf{S} \in \mathbb{R}^{K}$.

Without loss of generality, we define the range of each probing result in $\mathbb{R}$. Empirically, many scores are used as probing results, including correlation scores \citep{gupta-etal-2015-distributional}, usable $\mathcal{V}$-information \citep{pimentel-cotterell-2021-bayesian}, minimum description length \citep{voita-titov-2020-information}, and combinations thereof \citep{hewitt-liang-2019-designing}. Currently, the most popular probing results are written in accuracy values, ranging from 0 to 1. We do not require the actual choice of probing metric as long as all probing tasks adopt the same metric.

\paragraph{From model state to dataset state}
Now that we have the vectorized readout values of the states, how is the state $S$ determined? In general, there are three factors: the previous state, the dataset, and the training procedure applied to the model since the previous state. We can factor out the effects of the previous state and the training procedure.

To factor out the effect of the previous state, we introduce the concept of an ``initial state''. The initial state is specified by the structures and the parameters of the DNNs. If we study the effects of data in multitask fine-tuning BERT, then the initial state is the BERT model before any fine-tuning. Let us write the initial state as $S_0$. Based on this initial state, the dataset $X$ specifies a training task that leads the model from the initial state $S_0$ to the current state $S_{X}$. 

To factor out the effect of the training procedure, we assume the dataset $X$, together with the training objective, defines a unique global optimum. Additionally, we consider the training procedure can eventually reach this optimum, which we write as the \textit{dataset state}, $S_{X}$. 
Practically, various factors in the training procedures and the randomness in the sampled data may lead the models towards some local optima, but the framework considers the global optimum when specifying the dataset state.

\begin{figure}[t]
    \centering
    \includegraphics[width=\linewidth]{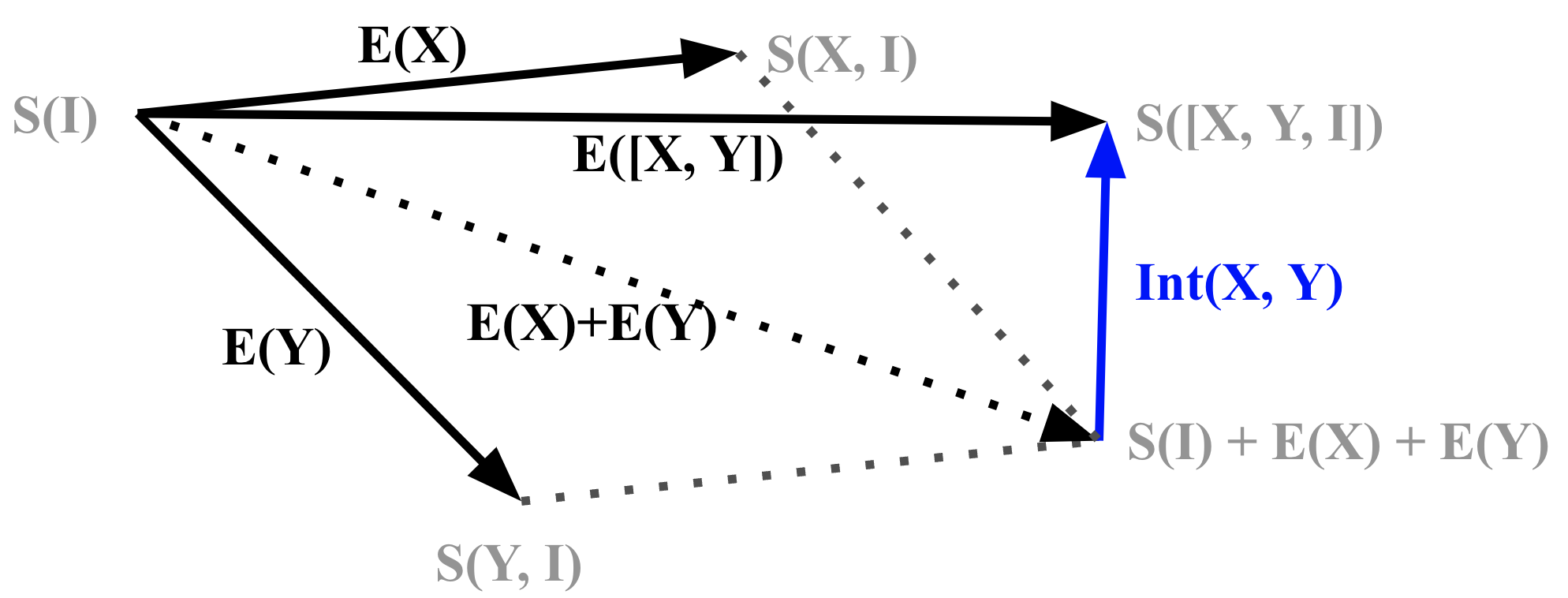}
    \caption{Schematic illustration of the dataset effects. $\mathbf{E(\cdot)}$ are the individual effects, $\mathbf{\textrm{Int}(X,Y)}$ is the interaction effect. }
    \label{fig:state_vectors}
\end{figure}

\section{Dataset effects}
\subsection{Individual dataset effect} 
We define the effect of a dataset $x$ as:
\begin{equation}
    \mathbf{E}(X) \equiv \mathbf{S}([X,I]) - \mathbf{S}(I) \in \mathbb{R}^{K}
    \label{eq:dataset-effect}
\end{equation}

Here $[X,I]$ denotes combining datasets $X$ and $I$, and $\mathbf{S}([X,I])$ and $\mathbf{S}(I)$ are the state vectors of the states $S_{[X,I]}$ and $S_I$ respectively. $\mathbf{E}(X)$ describes how does the linguistic ability of the model shift along the $K$ dimensions that we probe. In other words, $\mathbf{E}(X)$ describes the multi-dimensional fine-tuning effect of the dataset $X$.

\textit{Remark 1}: The effect of $X$ depends on a ``reference state'' $S_{I}$ where $I$ is used to describe the dataset leading to the reference state $S_I$. In an edge case, the dataset $X$ is a subset of the dataset $I$, so $S_{[X,I]}=S_{I}$. This can be attributed to the fact that our definition $S_{I}$ is the global optimum of a model fine-tuned on $I$ among all possible training procedures, including re-sampling. Hence, the dataset $X$ should have no effect on the model, which can be verified by $\mathbf{E}(X) = \mathbf{S}([X,I])-\mathbf{S}(I) = \mathbf{0}$. 

\textit{Remark 2}: In another scenario, $X$ consists of $I$ plus only one other data point $z$. Then $\mathbf{E}(X)$ degenerates to the effect of the data point $z$. 

\textit{Remark 3}: We assume $X$ does not overlap with $I$ in the rest of this paper. This assumption stands without loss of generality since we can always redefine $X$ as the non-overlapping data subset.

\textit{Remark 4}: The dataset effects form an Abelian group -- Appendix \S \ref{subsec:abelian-group-motivation} contains the details.

\subsection{Interaction effect} 
\paragraph{Motivating example} Let us first consider an example of detecting the sentiment polarity. Suppose three abilities can contribute to addressing the sentiment polarity task:
\begin{itemize}[nosep]
    \item[A1:] Parsing the structure of the review.
    \item[A2:] Recognizing the tense of the review.
    \item[A3:] Detecting some affective keywords such as ``good''.
\end{itemize}
Consider two sentiment polarity datasets, $X$ and $Y$. In $X$, all positive reviews follow a unique syntax structure where all negative reviews do not. In $Y$, all positive reviews are written in the present tense where all negative reviews in the past tense. The problem specified by dataset $X$ can be solved by relying on both A1 and A3, and the problem specified by dataset $Y$ can be solved by relying on both A2 and A3. 
Imagine a scenario where after training on both $X$ and $Y$, a model relies solely on A3 to predict the sentiment polarity. This behavior is caused by the interaction between $X$ and $Y$. Using the terms of our state-vector framework, there is an \textit{interaction effect} between $X$ and $Y$ with a positive value along the dimension of A3, and a negative value along the dimensions of A1 and A2.

\paragraph{Definition of the interaction effect}
Let us define the \textit{interaction effect} between two datasets, $X$ and $Y$ as:
\begin{equation}
    \textrm{Int}(X, Y) = \mathbf{E}([X,Y]) - (\mathbf{E}(X) + \mathbf{E}(Y))
    \label{eq:interaction-effect}
\end{equation}
This is the difference between the dataset effect of $[X,Y]$ ($X$ and $Y$ combined), and the sum of the effects of $X$ and $Y$ (as if $X$ and $Y$ have no interactions at all). 

\textit{Remark 1:} An equivalent formulation is:
\begin{equation}
\begin{aligned}
    \textrm{Int}(X,Y) = &\mathbf{S}([X,Y,I]) - \mathbf{S}([X,I]) \\ &- \mathbf{S}([Y,I]) + \mathbf{S}(I)
\end{aligned}
\label{eq:interaction-effect-equiv}
\end{equation}

\textit{Remark 2:} What if $X$ and $Y$ share some common items? This would introduce an additional effect when merging $X$ and $Y$. An extreme example is when $X$ and $Y$ are the same datasets, where $[X,Y]$ collapses to $X$. Then $\textrm{Int}(X,Y)=-\mathbf{E}(X)$ where it should be $\mathbf{0}$. The reason for this counter-intuitive observation is that the ``collapse'' step itself constitutes a significant interaction. Our ``datasets do not overlap'' assumption avoids this problem. This assumption goes without loss of generality because we can always redefine $X$ and $Y$ to contain distinct data points. 

\paragraph{A linear regression view}
The interaction effect as defined in Eq. \ref{eq:interaction-effect}, equals the $\beta_3$ coefficient computed by regressing for the state vector along each of the $K$ dimensions:
\begin{equation}
    \mathbf{S}^{(k)} = \beta_0^{(k)} + \beta_1^{(k)} i_x + \beta_2^{(k)} i_y + \beta_3^{(k)} i_xi_y + \epsilon 
    \label{eq:linreg-interaction}
\end{equation}
where $i_x$ and $i_y$ are indicator variables, and $\epsilon$ is the residual. If dataset $X$ appears, $i_x=1$, and $i_x=0$ otherwise. The same applies to $i_y$. $\mathbf{S}^{(k)}$ is the $k^{th}$ dimension in the state vector. The correspondence between the indicator variables and $\mathbf{S}$ as listed in Table \ref{tab:truth-values-regression}:

\begin{table}[h]
    \centering
    \begin{tabular}{l l l}
    \toprule 
        $i_x$ & $i_y$ & $\mathbf{S}$ \\ \midrule 
        0 & 0 & $\mathbf{S}(I)$ \\
        1 & 0 & $\mathbf{S}([X,I])$ \\
        0 & 1 & $\mathbf{S}([Y,I])$ \\
        1 & 1 & $\mathbf{S}([X,Y,I])$ \\ \bottomrule
    \end{tabular}
    \caption{Correspondence between the indicator variables and the regression targets.}
    \label{tab:truth-values-regression}
\end{table}

Please refer to Appendix \ref{subsec:interaction-effect-formulations-equivalence} for a derivation of the equivalence. The Eq.~\ref{eq:linreg-interaction} formulation allows us to apply an ANOVA, which allows us to decide if the interaction effect is significant (i.e., the $p$-value of $\beta_3$ is smaller than the significance threshold).

\section{Experiments}
\subsection{Models}
Experiments are conducted using two popular language models: BERT-base-cased \citep{devlin_BERT_2019} and RoBERTa-base \citep{liu_RoBERTa_2019}. Doing this allows us to compare results and determine whether dataset effects hold despite model choice. 

\subsection{Fine-tuning}
\label{section:finetune}
The GLUE benchmark \citep{wang-etal-2018-glue} consists of 3 types of natural language understanding tasks: single-sentence, similarity and paraphrase, and inference tasks. Two tasks from each category were selected to fine-tune models. 

\paragraph{Single-sentence tasks} 
COLA \citep{warstadt2018neural} labels whether a sentence is a grammatical English sentence or not. SST2 \citep{socher-etal-2013-recursive} labels whether a sentence has positive or negative sentiment. 

\paragraph{Similarity and paraphrase tasks}
MRPC \citep{dolan-brockett-2005-automatically} labels whether sentences in a pair are semantically equivalent or not. STSB \citep{cer-etal-2017-semeval} accomplishes the same objective except it provides a similarity score between 1-5.

\paragraph{Inference tasks} 
QNLI \citep{rajpurkar-etal-2016-squad} takes a sentence-question pair as input and labels whether the sentence contains the answer to the question. RTE \citep{dagan2005pascal,haim2006second,giampiccolo2007third,bentivogli2009fifth} labels whether a given conclusion is implied from some given text.

\paragraph{} 
The multitask model resembles that of \citet{radford_improving_2018}. It consists of a shared encoder with a separate classification layer per task -- Figure \ref{fig:finetune_setup} shows an illustration. This was made possible by leveraging HuggingFace's Transformers library \citep{wolf-etal-2020-transformers}. Model hyperparameters are the same as those listed in Table 3 of \citet{mosbach-etal-2020-interplay-fine}, except 3 epochs were used for all experiments. This is valid, as their experiments incorporated both BERT-base-cased and RoBERTa-base models. Model checkpoints were saved every 6 optimization steps, with the final model being the one with the lowest training loss.\footnote{Downstream task performance can be accessed in our repository.}

To mitigate the effect of train set size per task, train datasets were reduced to 2,490 examples per task, which corresponds to the size of the smallest train dataset (COLA). GLUE benchmarks such as QQP, MNLI, and WNLI \citep{levesque2012winograd} were excluded because their train sets were too large (more than 300K examples) or too small (fewer than 1K examples). 

\subsection{Probing}
We use the SentEval suite \citep{conneau-kiela-2018-senteval} to build proxies for the idealized probes that vectorize the model states. SentEval contains the following tasks:

\begin{itemize}[nosep]
    \item \textbf{Length:} given a sentence, predict what range its length falls within (0: 5-8, 1: 9-12, 2: 13-16, 3: 17-20, 4: 21-25, 5: 26-28). 
    \item \textbf{WC:} given a sentence, predict which word it contains from a target set of 1,000 words.
    \item \textbf{Depth:} given a sentence, predict the maximum depth of its syntactic tree.
    \item \textbf{TopConst:} given a sentence, predict its constituent sequence (e.g. NP\_VP\_.: noun phrase followed by verb phrase).
    \item \textbf{BigramShift:} given a sentence, predict whether any two consecutive tokens of the original sentence have been inverted.
    \item \textbf{Tense:} given a sentence, predict whether its main verb is in the past or present tense.
    \item \textbf{SubjNumber:} given a sentence, predict whether the subject of its main clause is singular or plural.
    \item \textbf{ObjNumber:} given a sentence, predict whether the direct object of its main clause is singular or plural.
    \item \textbf{OddManOut:} given a sentence, predict whether any verb or noun of the original sentence was replaced with another form with the same part of speech.
    \item \textbf{CoordInv:} given a sentence, predict whether two coordinated casual conjoints of the original sentence have been inverted.
\end{itemize}

\paragraph{}
For each probing task, we downsample the datasets to 10\% of their original sizes or 12K samples per task (10K train, 1K validation, 1K test). This is valid, as datasets of similar sizes usually have sufficient statistical power \citep{card_little_2020,zhu_data_2022}. WC was removed from consideration, as its performance would have been significantly compromised given that it possesses 1000 ground truths. 
We built our training pipeline based on the SentEval library and used the default config.\footnote{SentEval code and default config are accessible \href{https://github.com/facebookresearch/SentEval}{here}.} The default architecture is a single classification layer on top of the frozen fine-tuned encoder.

\subsection{Multitask settings}
As the number of tasks increases, the number of combinations increases exponentially. To compute the dataset states in a realistic time, we group the experiments by the format of the tasks (e.g., single-sentence, similarity, inference) and impose selection conditions, reducing the total number of fine-tuning experiments from around 10k to 300 per model. Note that 300 fine-tuning experiments per model are still nontrivial, but they can be completed within two months. Section \ref{subsec:appendix-multitask-settings} in Appendix contains the details.

\section{Results}
\begin{figure}[t]
    \centering
    \includegraphics[width=\linewidth]{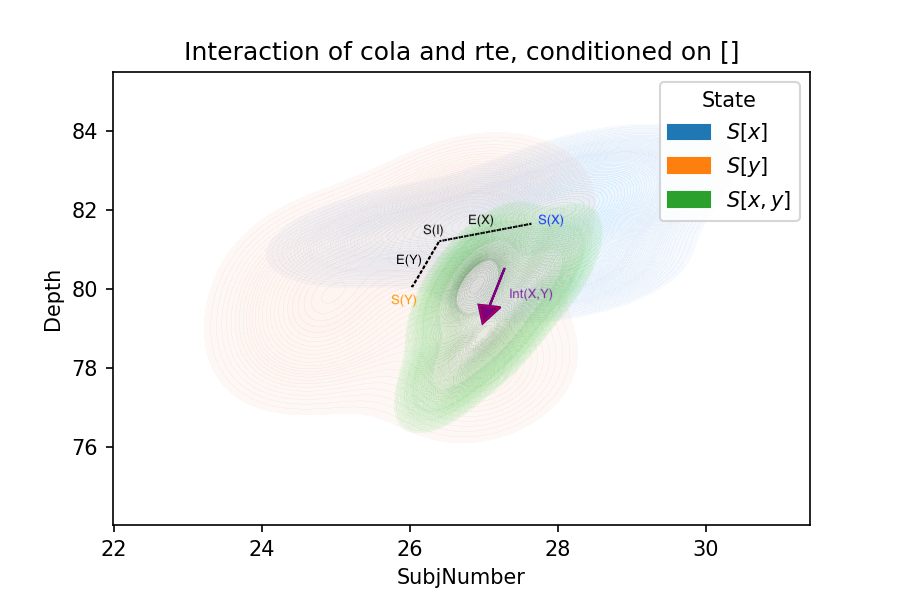}
    \caption{An example of interaction effect between x=COLA and y=RTE, of RoBERTa. The purple arrow visualizes the interaction effect. This figure plots two linguistic ability dimensions -- syntax tree depth and subject number.}
    \label{fig:interaction_effect_example}
\end{figure}

\begin{table*}    
\resizebox{\linewidth}{!}{
\begin{tabular}{lllllllllll}
\toprule
Dataset & Model &                   Length &                       Depth &                 TopConst &                 BigramShift &         Tense &                SubjNumber &                   ObjNumber &                   OddManOut &                    CoordInv \\
\midrule
   COLA &  BERT &               $-0.36^{}$ &                  $-0.47^{}$ &               $-0.83^{}$ & $\hspace{0.75em}6.35^{***}$ &    $-0.23^{}$ &                $-0.38^{}$ &                 $-0.69^{*}$ & $\hspace{0.75em}1.93^{***}$ & $\hspace{0.75em}1.09^{***}$ \\
   SST2 &  BERT &            $-4.08^{***}$ &                  $-0.51^{}$ &            $-3.92^{***}$ &               $-1.99^{***}$ &    $-0.24^{}$ &             $-1.75^{***}$ &               $-1.73^{***}$ &                 $-0.79^{*}$ &    $\hspace{0.75em}0.06^{}$ \\
   MRPC &  BERT & $\hspace{0.75em}0.65^{}$ & $\hspace{0.75em}0.75^{***}$ &            $-2.34^{***}$ &    $\hspace{0.75em}0.55^{}$ & $-0.45^{***}$ &               $-0.78^{*}$ & $\hspace{0.75em}1.07^{***}$ & $\hspace{0.75em}1.13^{***}$ & $\hspace{0.75em}1.34^{***}$ \\
   STSB &  BERT &               $-0.53^{}$ &    $\hspace{0.75em}0.27^{}$ &              $-1.66^{*}$ &                  $-0.24^{}$ &    $-0.3^{*}$ &  $\hspace{0.75em}0.36^{}$ &     $\hspace{0.75em}0.2^{}$ &    $\hspace{0.75em}0.36^{}$ &                  $-0.09^{}$ \\
   QNLI &  BERT & $\hspace{0.75em}0.05^{}$ &    $\hspace{0.75em}0.33^{}$ & $\hspace{0.75em}0.85^{}$ &                  $-0.49^{}$ &    $-0.16^{}$ & $\hspace{0.75em}0.81^{*}$ &    $\hspace{0.75em}0.16^{}$ &                  $-0.34^{}$ & $\hspace{0.75em}1.33^{***}$ \\
    RTE &  BERT & $\hspace{0.75em}0.83^{}$ &    $\hspace{0.75em}0.34^{}$ &               $-0.58^{}$ &                  $-0.24^{}$ &    $-0.21^{}$ &                $-0.15^{}$ &                 $-0.65^{*}$ &    $\hspace{0.75em}0.16^{}$ &    $\hspace{0.75em}0.64^{}$ \\
\bottomrule
\end{tabular}
}
\caption{Individual effects of GLUE datasets, on BERT. $^{*}$, $^{**}$, and $^{***}$ stand for $p<0.05$, $p<0.01$, and $p<0.001$, respectively, using two-sample $t$-test with $dof=8$.}
\label{tab:ind-effects-BERT}
\end{table*}

\begin{table*}    
\resizebox{\linewidth}{!}{
\begin{tabular}{lllllllllll}
\toprule
Dataset &   Model &                   Length &                    Depth &                  TopConst &                 BigramShift &                       Tense &               SubjNumber &                ObjNumber &                   OddManOut &                   CoordInv \\
\midrule
   COLA & RoBERTa &            $-2.75^{***}$ & $\hspace{0.75em}0.28^{}$ & $\hspace{0.75em}2.17^{*}$ & $\hspace{0.75em}9.17^{***}$ & $\hspace{0.75em}1.21^{***}$ & $\hspace{0.75em}0.26^{}$ & $\hspace{0.75em}0.59^{}$ & $\hspace{0.75em}3.66^{***}$ & $\hspace{0.75em}2.0^{***}$ \\
   SST2 & RoBERTa &            $-7.12^{***}$ &            $-1.86^{***}$ &             $-5.88^{***}$ &               $-4.43^{***}$ &               $-1.57^{***}$ &            $-3.92^{***}$ &            $-3.96^{***}$ &               $-1.85^{***}$ &              $-2.08^{***}$ \\
   MRPC & RoBERTa &               $-0.06^{}$ & $\hspace{0.75em}0.25^{}$ &                $-0.87^{}$ &                  $-1.68^{}$ &                 $-1.01^{*}$ &               $-0.61^{}$ &               $-0.45^{}$ &    $\hspace{0.75em}0.07^{}$ &    $\hspace{0.75em}1.0^{}$ \\
   STSB & RoBERTa & $\hspace{0.75em}0.03^{}$ & $\hspace{0.75em}0.14^{}$ &             $-3.17^{***}$ &                  $-3.0^{*}$ &               $-1.84^{***}$ &            $-3.11^{***}$ &              $-1.65^{*}$ &                  $-0.43^{}$ &   $\hspace{0.75em}0.16^{}$ \\
   QNLI & RoBERTa &               $-0.67^{}$ & $\hspace{0.75em}0.04^{}$ &                $-0.13^{}$ &                  $-0.57^{}$ &                  $-0.56^{}$ &               $-0.33^{}$ &              $-1.47^{*}$ &     $\hspace{0.75em}0.5^{}$ &   $\hspace{0.75em}0.13^{}$ \\
    RTE & RoBERTa & $\hspace{0.75em}0.78^{}$ &              $-0.84^{*}$ &                $-0.58^{}$ &                  $-0.32^{}$ &                  $-0.81^{}$ &               $-0.85^{}$ &            $-2.04^{***}$ &                  $-0.25^{}$ &                   $\hspace{0.75em}0.0^{}$ \\
\bottomrule
\end{tabular}
}
\caption{Individual effects of GLUE datasets, on RoBERTa. $^{*}$, $^{**}$, and $^{***}$ stand for $p<0.05$, $p<0.01$, and $p<0.001$, respectively, using two-sample $t$-tests with $dof=8$.}
\label{tab:ind-effects-RoBERTa}
\end{table*}

\subsection{Individual dataset effect}
Some observations regarding individual dataset effects are noted below. Tables \ref{tab:ind-effects-BERT} and \ref{tab:ind-effects-RoBERTa} summarize average individual GLUE dataset effects for BERT and RoBERTa models, respectively. Tables \ref{tab:ind-effects-cola} and \ref{tab:ind-effects-sst2} break down the average individual effects of COLA and SST2, respectively.\footnote{Tables for other datasets can be accessed in our repository.}

\paragraph{Model choice}
Per Tables \ref{tab:ind-effects-BERT} and  \ref{tab:ind-effects-RoBERTa}, there is no consistent agreement between significant dimensions (i.e., dimensions marked as $^{*}$, $^{**}$, or $^{***}$). In fact, of the combined 33 significant individual effects observed in Tables \ref{tab:ind-effects-BERT} and \ref{tab:ind-effects-RoBERTa}, only 13 ($\approx 40\%$) can be confirmed by both BERT and RoBERTa models. This demonstrates that datasets can have different effects depending on model architecture. 

\paragraph{Dataset composition} 
MRPC and STSB accomplish very similar tasks, but impact different dimensions. Matching significant dimensions for MRPC and STSB amount to 2 of 7 for BERT (see Table \ref{tab:ind-effects-BERT}) and 1 of 5 for RoBERTa (see Table \ref{tab:ind-effects-RoBERTa}). Although they both are paraphrasing tasks, their samples are extracted from different sources and have different ground truths (i.e., MRPC is a binary task, STSB is an ordinal task). Hence, the composition of datasets can affect what individual effects they will contribute. 

\paragraph{Dataset type} 
The inference datasets (QNLI, RTE) do not have much significant impact on the set of probing dimensions. In all cases, both datasets have no more than two significant dimensions (see Tables \ref{tab:ind-effects-BERT} and \ref{tab:ind-effects-RoBERTa}). This is far fewer than single-sentence tasks (COLA, SST2), whose effects span many more dimensions (see Tables \ref{tab:ind-effects-BERT} and \ref{tab:ind-effects-RoBERTa}). We hypothesize that this can be attributed to the fact that SentEval probing tasks assess linguistic information captured on the sentence level. By fine-tuning on single-sentence datasets, models are more likely to learn these relevant sentence properties and incorporate them in their embeddings. Inference datasets are more complex and require models to learn linguistic properties beyond the sentence scope, such as semantic relationships between sentences. Similarity and paraphrase datasets fall in between single-sentence and inference datasets with respect to complexity and linguistic scope, which explains why MPRC and STSB impact more dimensions than QNLI and RTE but fewer dimensions than COLA and SST2. 

\begin{table*}    
\resizebox{\linewidth}{!}{
\begin{tabular}{llllllllllll}
\toprule
      Dataset & Reference & Model &                   Length &                    Depth &                 TopConst &                 BigramShift &                    Tense &               SubjNumber &   ObjNumber &                   OddManOut &                    CoordInv \\
\midrule
          COLA & $I$ &  BERT &               $-1.32^{}$ &  $\hspace{0.75em}0.8^{}$ &              $-2.94^{*}$ & $\hspace{0.75em}6.32^{***}$ &            $-1.14^{***}$ &               $-0.02^{}$ &  $-0.34^{}$ & $\hspace{0.75em}3.56^{***}$ &     $\hspace{0.75em}0.3^{}$ \\
     COLA & SST2 &  BERT &               $-1.16^{}$ &                $-1.2^{}$ &                $-0.0^{}$ &  $\hspace{0.75em}7.4^{***}$ &                $-0.7^{}$ & $\hspace{0.75em}0.22^{}$ &   $-0.7^{}$ & $\hspace{0.75em}2.22^{***}$ &    $\hspace{0.75em}1.82^{}$ \\
     COLA & MRPC &  BERT &                $-0.6^{}$ & $\hspace{0.75em}0.78^{}$ &               $-0.06^{}$ &  $\hspace{0.75em}5.6^{***}$ & $\hspace{0.75em}0.44^{}$ & $\hspace{0.75em}0.14^{}$ &  $-0.18^{}$ &    $\hspace{0.75em}0.56^{}$ &    $\hspace{0.75em}0.92^{}$ \\
     COLA & STSB &  BERT &               $-2.84^{}$ &                $-1.0^{}$ &                $-0.8^{}$ & $\hspace{0.75em}6.48^{***}$ & $\hspace{0.75em}0.94^{}$ & $\hspace{0.75em}0.08^{}$ & $-1.94^{*}$ &   $\hspace{0.75em}3.34^{*}$ & $\hspace{0.75em}3.54^{***}$ \\
     COLA & QNLI &  BERT & $\hspace{0.75em}1.24^{}$ &               $-1.26^{}$ &               $-1.26^{}$ &  $\hspace{0.75em}7.2^{***}$ &               $-0.24^{}$ &               $-0.96^{}$ &  $-0.66^{}$ &    $\hspace{0.75em}2.46^{}$ &    $\hspace{0.75em}0.58^{}$ \\
      COLA & RTE &  BERT &  $\hspace{0.75em}1.5^{}$ &              $-1.82^{*}$ &               $-3.3^{*}$ & $\hspace{0.75em}5.94^{***}$ &               $-0.26^{}$ &               $-2.02^{}$ &  $-0.68^{}$ &    $\hspace{0.75em}2.16^{}$ &    $\hspace{0.75em}0.96^{}$ \\
COLA & MRPC QNLI &  BERT & $\hspace{0.75em}0.62^{}$ & $\hspace{0.75em}0.46^{}$ &               $-2.86^{}$ & $\hspace{0.75em}5.22^{***}$ &              $-1.08^{*}$ &               $-0.62^{}$ &  $-0.46^{}$ &    $\hspace{0.75em}1.12^{}$ &    $\hspace{0.75em}0.28^{}$ \\
 COLA & MRPC RTE &  BERT &               $-0.58^{}$ &               $-0.64^{}$ & $\hspace{0.75em}0.92^{}$ & $\hspace{0.75em}5.52^{***}$ &               $-0.32^{}$ & $\hspace{0.75em}1.34^{}$ &  $-0.24^{}$ & $\hspace{0.75em}1.46^{***}$ &    $\hspace{0.75em}1.02^{}$ \\
COLA & STSB QNLI &  BERT &               $-0.32^{}$ &               $-0.52^{}$ & $\hspace{0.75em}1.12^{}$ & $\hspace{0.75em}6.56^{***}$ & $\hspace{0.75em}0.24^{}$ &               $-1.26^{}$ &  $-0.44^{}$ &   $\hspace{0.75em}1.38^{*}$ &    $\hspace{0.75em}0.38^{}$ \\
 COLA & STSB RTE &  BERT &               $-0.18^{}$ &               $-0.26^{}$ & $\hspace{0.75em}0.92^{}$ & $\hspace{0.75em}7.24^{***}$ &                $-0.2^{}$ &                $-0.7^{}$ &  $-1.22^{}$ &    $\hspace{0.75em}1.08^{}$ &    $\hspace{0.75em}1.08^{}$ \\ \midrule 
          COLA & $I$ & RoBERTa & $\hspace{0.75em}2.18^{***}$ &  $\hspace{0.75em}1.24^{}$ & $\hspace{0.75em}8.62^{***}$ &  $\hspace{0.75em}5.78^{***}$ &  $\hspace{0.75em}0.28^{}$ & $\hspace{0.75em}0.44^{}$ &              $-1.78^{*}$ & $\hspace{0.75em}4.64^{***}$ &    $\hspace{0.75em}2.7^{*}$ \\
     COLA & SST2 & RoBERTa &                  $-0.66^{}$ &  $\hspace{0.75em}2.18^{}$ & $\hspace{0.75em}6.18^{***}$ & $\hspace{0.75em}13.08^{***}$ &  $\hspace{0.75em}2.44^{}$ & $\hspace{0.75em}3.28^{}$ & $\hspace{0.75em}2.42^{}$ &  $\hspace{0.75em}4.4^{***}$ &   $\hspace{0.75em}4.52^{*}$ \\
     COLA & MRPC & RoBERTa &                 $-5.48^{*}$ &                $-0.34^{}$ &                  $-1.58^{}$ &  $\hspace{0.75em}7.94^{***}$ &  $\hspace{0.75em}0.14^{}$ & $\hspace{0.75em}0.16^{}$ & $\hspace{0.75em}0.78^{}$ &   $\hspace{0.75em}2.62^{*}$ &     $\hspace{0.75em}1.6^{}$ \\
     COLA & STSB & RoBERTa &                 $-3.84^{*}$ &                 $-1.0^{}$ &     $\hspace{0.75em}3.8^{}$ &  $\hspace{0.75em}11.9^{***}$ & $\hspace{0.75em}3.46^{*}$ & $\hspace{0.75em}0.98^{}$ &                $-0.1^{}$ &    $\hspace{0.75em}4.6^{*}$ &     $\hspace{0.75em}2.2^{}$ \\
     COLA & QNLI & RoBERTa &                  $-3.68^{}$ &                 $-0.5^{}$ &    $\hspace{0.75em}0.38^{}$ &  $\hspace{0.75em}5.86^{***}$ &                 $-0.1^{}$ &               $-1.88^{}$ &               $-0.96^{}$ & $\hspace{0.75em}3.84^{***}$ &   $\hspace{0.75em}3.02^{*}$ \\
      COLA & RTE & RoBERTa &                  $-1.82^{}$ &  $\hspace{0.75em}1.04^{}$ &   $\hspace{0.75em}6.04^{*}$ &  $\hspace{0.75em}7.78^{***}$ &  $\hspace{0.75em}0.36^{}$ &                $-0.5^{}$ & $\hspace{0.75em}0.28^{}$ & $\hspace{0.75em}4.18^{***}$ & $\hspace{0.75em}4.28^{***}$ \\
COLA & MRPC QNLI & RoBERTa &                 $-5.58^{*}$ &                 $-0.7^{}$ &    $\hspace{0.75em}0.24^{}$ &   $\hspace{0.75em}8.8^{***}$ &  $\hspace{0.75em}0.78^{}$ &  $\hspace{0.75em}1.0^{}$ & $\hspace{0.75em}0.86^{}$ &   $\hspace{0.75em}2.18^{*}$ &    $\hspace{0.75em}0.54^{}$ \\
 COLA & MRPC RTE & RoBERTa &    $\hspace{0.75em}0.56^{}$ & $\hspace{0.75em}1.34^{*}$ &     $\hspace{0.75em}3.1^{}$ &  $\hspace{0.75em}8.64^{***}$ & $\hspace{0.75em}3.32^{*}$ & $\hspace{0.75em}1.54^{}$ & $\hspace{0.75em}2.54^{}$ &    $\hspace{0.75em}3.3^{*}$ &    $\hspace{0.75em}0.62^{}$ \\
COLA & STSB QNLI & RoBERTa &                  $-4.04^{}$ &  $\hspace{0.75em}0.38^{}$ &    $\hspace{0.75em}0.88^{}$ & $\hspace{0.75em}11.02^{***}$ &  $\hspace{0.75em}1.22^{}$ & $\hspace{0.75em}0.28^{}$ & $\hspace{0.75em}1.28^{}$ &  $\hspace{0.75em}3.4^{***}$ &     $\hspace{0.75em}0.1^{}$ \\
 COLA & STSB RTE & RoBERTa &               $-5.12^{***}$ &                 $-0.8^{}$ &                  $-5.98^{}$ & $\hspace{0.75em}10.86^{***}$ &  $\hspace{0.75em}0.22^{}$ &               $-2.74^{}$ & $\hspace{0.75em}0.62^{}$ &    $\hspace{0.75em}3.46^{}$ &    $\hspace{0.75em}0.42^{}$ \\
\bottomrule
\end{tabular}
}
\caption{Individual effects of COLA dataset with different reference states. $^{*}$, $^{**}$, and $^{***}$ stand for $p<0.05$, $p<0.01$, and $p<0.001$, respectively, computed using two-sample $t$-test with $dof=8$.}
\label{tab:ind-effects-cola}
\end{table*}

\paragraph{Reference state}
In most cases, significant dataset dimensions varied with different reference states. From Table \ref{tab:ind-effects-cola}, it is clear that significant individual effects of COLA are inconsistent between experiments, besides  BigramShift (for both models) and OddManOut (positive effect for RoBERTa only). The same conclusion is valid for datasets other than COLA. This implies that there are inherent interaction effects between datasets that also influence results. Note that if we add a dataset to a large number of varying reference states and observe that there are persistent, significant dimensions across these experiments, then this is a strong empirical indication of a dataset's effect on this set of dimensions (e.g., see COLA's effect on BigramShift in Table \ref{tab:ind-effects-cola} and SST2's effect on Length in Table \ref{tab:ind-effects-sst2}). In the case of our experiments, for argument's sake, we impose that a dataset effect must appear in at least 70\% reference states. This lower bound can be adjusted to 60\% or 80\%, but wouldn't result in any major adjustments as dataset effects tend to appear in fewer than 50\% of our experiments. Table \ref{tab:sig-ind-effects} summarizes such instances and supports previous statements made regarding model choice (some effects are only observed on RoBERTa) and dataset type (both datasets are single-sentence tasks). The low number of table entries further justifies that there are other confounding variables, including but not limited to model choice, hyperparameter selection, and dataset interactions. 

\begin{table}[h]
    \centering
    \resizebox{0.9\linewidth}{!}{
    \begin{tabular}{cccc}
    \toprule 
        Dataset & Dimension & Effect & Model(s) \\ \midrule 
        COLA & BigramShift & + & Both \\
        COLA & OddManOut & + & RoBERTa \\
        SST2 & Length & - & Both \\
        SST2 & TopConst & - & RoBERTa \\
        SST2 & BigramShift & - & Both \\ \bottomrule
    \end{tabular}}
    \caption{Individual effects that are observed in at least 70\% of reference states.}
    \label{tab:sig-ind-effects}
\end{table} 

\paragraph{Spill-over}
We observed that some syntactic datasets have effects along the semantic dimensions and vice-versa. This is odd, as learning sentiment shouldn't be correlated to e.g., losing the ability to identify the syntactic constituents or swapped word order. We refer to such effects as ``spill-over'' effects. For instance, Tables \ref{tab:ind-effects-BERT} and \ref{tab:ind-effects-RoBERTa} suggest that fine-tuning on COLA (a syntactic dataset) has a positive effect on OddManOut and CoordInv (semantic dimensions). This is unexpected, given OddManOut and CoordInv probing datasets consist of grammatically acceptable sentences, they only violate semantic rules. Conversely, fine-tuning on SST2 (a semantic dataset) hurts TopConst and BigramShift (syntactic or surface-form dimensions). We hypothesize that the spill-over individual effects are likely due to the aforementioned confounding variables inducing false correlations. More rigorous analysis is required to identify these variables and their effects. 

\begin{table*}    
\resizebox{\linewidth}{!}{
\begin{tabular}{lll | lllllllll}
\toprule
 \multirow{2}{*}{$X$} & \multirow{2}{*}{$Y$} &      \multirow{2}{*}{Model} & \multicolumn{9}{c}{Dataset effect dimensions} \\
 & & &                    Length &                    Depth &            TopConst &                 BigramShift &                      Tense &                 SubjNumber &                  ObjNumber &                OddManOut &       CoordInv \\
\midrule
 COLA &    SST2 & RoBERTa &               $-2.84^{}$ & $\hspace{0.75em}0.94^{}$ &               $-2.44^{}$ & $\hspace{0.75em}7.30^{***}$ &   $\hspace{0.75em}2.16^{}$ & $\hspace{0.75em}2.84^{}$ & $\hspace{0.75em}4.20^{*}$ &   $-0.24^{}$ &  $\hspace{0.75em}1.82^{}$ \\
   COLA &    MRPC & RoBERTa &            $-7.66^{***}$ &               $-1.58^{}$ &           $-10.20^{***}$ &  $\hspace{0.75em}2.16^{**}$ &                 $-0.14^{}$ &               $-0.28^{}$ &  $\hspace{0.75em}2.56^{}$ &   $-2.02^{}$ &                $-1.10^{}$ \\
   COLA &    STSB & RoBERTa &             $-6.02^{**}$ &              $-2.24^{*}$ &               $-4.82^{}$ & $\hspace{0.75em}6.12^{***}$ & $\hspace{0.75em}3.18^{**}$ & $\hspace{0.75em}0.54^{}$ &  $\hspace{0.75em}1.68^{}$ &   $-0.04^{}$ &                $-0.50^{}$ \\
   COLA &    QNLI & RoBERTa &             $-5.86^{**}$ &              $-1.74^{*}$ &            $-8.24^{***}$ &    $\hspace{0.75em}0.08^{}$ &                 $-0.38^{}$ &               $-2.32^{}$ &  $\hspace{0.75em}0.82^{}$ &   $-0.80^{}$ &  $\hspace{0.75em}0.32^{}$ \\
   COLA &     RTE & RoBERTa &              $-4.00^{*}$ &               $-0.20^{}$ &               $-2.58^{}$ &  $\hspace{0.75em}2.00^{**}$ &   $\hspace{0.75em}0.08^{}$ &               $-0.94^{}$ &  $\hspace{0.75em}2.06^{}$ &   $-0.46^{}$ &  $\hspace{0.75em}1.58^{}$ \\
   \midrule 
   SST2 &    MRPC & RoBERTa &              $-3.74^{*}$ & $\hspace{0.75em}0.96^{}$ &                 $-2.72^{}$ &   $\hspace{0.75em}5.64^{*}$ &   $\hspace{0.75em}1.98^{}$ &  $\hspace{0.75em}4.40^{*}$ &  $\hspace{0.75em}3.74^{*}$ & $\hspace{0.75em}0.88^{}$ &                  $-0.50^{}$ \\
   SST2 &    STSB & RoBERTa &               $-2.86^{}$ & $\hspace{0.75em}0.06^{}$ &   $\hspace{0.75em}0.36^{}$ & $\hspace{0.75em}6.64^{***}$ &  $\hspace{0.75em}4.40^{*}$ &   $\hspace{0.75em}4.20^{}$ &  $\hspace{0.75em}3.60^{*}$ & $\hspace{0.75em}1.44^{}$ &    $\hspace{0.75em}1.44^{}$ \\
   SST2 &    QNLI & RoBERTa & $\hspace{0.75em}0.18^{}$ & $\hspace{0.75em}1.02^{}$ &   $\hspace{0.75em}0.18^{}$ &    $\hspace{0.75em}2.12^{}$ &   $\hspace{0.75em}2.64^{}$ &   $\hspace{0.75em}3.08^{}$ &   $\hspace{0.75em}3.84^{}$ & $\hspace{0.75em}0.60^{}$ &   $\hspace{0.75em}2.82^{*}$ \\
   SST2 &     RTE & RoBERTa &               $-0.20^{}$ & $\hspace{0.75em}1.48^{}$ & $\hspace{0.75em}8.14^{**}$ & $\hspace{0.75em}5.78^{***}$ & $\hspace{0.75em}4.68^{**}$ & $\hspace{0.75em}5.56^{**}$ & $\hspace{0.75em}5.48^{**}$ & $\hspace{0.75em}0.50^{}$ & $\hspace{0.75em}5.96^{***}$ \\
   \midrule
   
   MRPC &    STSB & RoBERTa &               $-3.34^{}$ &             $-3.14^{**}$ &                 $-1.08^{}$ &  $\hspace{0.75em}3.84^{*}$ & $\hspace{0.75em}3.58^{**}$ &  $\hspace{0.75em}2.98^{}$ &  $\hspace{0.75em}0.50^{}$ &               $-1.92^{}$ &                 $-0.78^{}$ \\
   MRPC &    QNLI & RoBERTa &               $-2.04^{}$ &               $-1.14^{}$ &                 $-4.44^{}$ &   $\hspace{0.75em}0.64^{}$ &   $\hspace{0.75em}0.42^{}$ &  $\hspace{0.75em}0.32^{}$ &  $\hspace{0.75em}1.44^{}$ & $\hspace{0.75em}0.90^{}$ &   $\hspace{0.75em}1.06^{}$ \\
   MRPC &     RTE & RoBERTa &             $-6.58^{**}$ &              $-1.58^{*}$ &                 $-3.10^{}$ &   $\hspace{0.75em}1.88^{}$ &                 $-1.48^{}$ &  $\hspace{0.75em}0.78^{}$ &  $\hspace{0.75em}0.04^{}$ &               $-0.10^{}$ &   $\hspace{0.75em}1.72^{}$ \\

   \midrule
   STSB &    QNLI & RoBERTa & $\hspace{0.75em}0.12^{}$ & $-2.42^{*}$ &                $-3.10^{}$ &  $\hspace{0.75em}1.18^{}$ &  $\hspace{0.75em}2.78^{}$ & $\hspace{0.75em}0.54^{}$ & $\hspace{0.75em}0.78^{}$ & $\hspace{0.75em}0.14^{}$ &  $\hspace{0.75em}1.16^{}$ \\
   STSB &     RTE & RoBERTa & $\hspace{0.75em}1.12^{}$ &  $-1.42^{}$ & $\hspace{0.75em}6.28^{*}$ & $\hspace{0.75em}3.28^{*}$ &  $\hspace{0.75em}3.12^{}$ & $\hspace{0.75em}1.52^{}$ & $\hspace{0.75em}1.84^{}$ & $\hspace{0.75em}0.26^{}$ &  $\hspace{0.75em}2.00^{}$ \\
   \midrule
   QNLI &     RTE & RoBERTa &               $-0.10^{}$ & $-0.74^{}$ &      $-0.68^{}$ & $\hspace{0.75em}0.96^{}$ & $\hspace{0.75em}0.06^{}$ & $-0.76^{}$ & $\hspace{0.75em}1.30^{}$ & $\hspace{0.75em}0.40^{}$ & $\hspace{0.75em}1.86^{}$ \\
\bottomrule
\end{tabular}
}
\caption{Interaction effects between GLUE datasets, on RoBERTa. $^{*}$, $^{**}$ and $^{***}$ stand for $p<0.05$, $p<0.01$, and $p<0.001$, respectively, from the ANOVA test of the interaction effect.}
\label{tab:interaction_effects_RoBERTa}
\end{table*}

\subsection{Interaction effect}
The average interaction effects of datasets on RoBERTa and BERT are listed in Table \ref{tab:interaction_effects_RoBERTa} and Table \ref{tab:interaction_effects_BERT}, respectively.

\paragraph{Interaction effects are concentrated} The interaction effects are not always significant along the probing dimensions. More specifically, in most (28 out of 30) scenarios listed in Tables \ref{tab:interaction_effects_RoBERTa} and Table \ref{tab:interaction_effects_BERT}, the significant interactions concentrate on no more than three dimensions. The remaining two scenarios are both SST2 and RTE (on RoBERTa and BERT, respectively).

\paragraph{Interaction effects can occur with insignificant individual effects} Many interaction effects are observed along the linguistic dimensions where neither of the participating datasets has a significant individual effect. For example, all significant interactions along the ``Depth'' dimension for BERT have this characteristic. Apparently, even if a dataset does not have a significant individual effect along a dimension, it could still interact with the other dataset along this dimension.

\paragraph{Interaction effects are characteristic} The significant dimensions differ across the datasets. Even the datasets with similar purposes do not demonstrate identical interaction effects with the same ``third-party'' dataset. For example, both MRPC and STSB target at detecting paraphrasing. When interacting with COLA, STSB has an insignificant effect on the TopConst dimension, while MRPC has a significant negative effect along the same dimension. Can we predict the dimensions of the significant effects, just by analyzing the datasets? Rigorous, systematic studies in the future are necessary to give deterministic answers.

\paragraph{Similar datasets interact less} The interactions of similar datasets appear to interact less significantly than those ``less similar'' datasets. 
Among the 30 scenarios in Tables \ref{tab:interaction_effects_RoBERTa} and \ref{tab:interaction_effects_BERT}, only two scenarios show no significant interactions along any dimensions: (MRPC, QNLI) and (QNLI, RTE), both on RoBERTa, and both involve strong similarities between the datasets: QNLI and RTE test the same downstream task (infer the textual entailment), and MRPC and QNLI both involve an intricate understanding of the semantics of the text.

\begin{table*}    
\resizebox{\linewidth}{!}{
\begin{tabular}{lll | lllllllll}
\toprule
 \multirow{2}{*}{$X$} & \multirow{2}{*}{$Y$} &      \multirow{2}{*}{Model} & \multicolumn{9}{c}{Dataset effect dimensions} \\
 & & &                    Length &                    Depth &            TopConst &                 BigramShift &                      Tense &                 SubjNumber &                  ObjNumber &                OddManOut &       CoordInv \\
\midrule
COLA &    SST2 &    BERT & $\hspace{0.75em}0.16^{}$ &              $-2.00^{*}$ & $\hspace{0.75em}2.94^{}$ &   $\hspace{0.75em}1.08^{*}$ &   $\hspace{0.75em}0.44^{}$ & $\hspace{0.75em}0.24^{}$ &                $-0.36^{}$ &   $-1.34^{}$ &  $\hspace{0.75em}1.52^{}$ \\
   COLA &    MRPC &    BERT & $\hspace{0.75em}0.72^{}$ &               $-0.02^{}$ & $\hspace{0.75em}2.88^{}$ &                  $-0.72^{}$ &  $\hspace{0.75em}1.58^{*}$ & $\hspace{0.75em}0.16^{}$ &  $\hspace{0.75em}0.16^{}$ & $-3.00^{**}$ &  $\hspace{0.75em}0.62^{}$ \\
   COLA &    STSB &    BERT &               $-1.52^{}$ &               $-1.80^{}$ & $\hspace{0.75em}2.14^{}$ &    $\hspace{0.75em}0.16^{}$ & $\hspace{0.75em}2.08^{**}$ & $\hspace{0.75em}0.10^{}$ &                $-1.60^{}$ &   $-0.22^{}$ & $\hspace{0.75em}3.24^{*}$ \\
   COLA &    QNLI &    BERT & $\hspace{0.75em}2.56^{}$ &              $-2.06^{*}$ & $\hspace{0.75em}1.68^{}$ &    $\hspace{0.75em}0.88^{}$ &  $\hspace{0.75em}0.90^{*}$ &               $-0.94^{}$ &                $-0.32^{}$ &   $-1.10^{}$ &  $\hspace{0.75em}0.28^{}$ \\
   COLA &     RTE &    BERT & $\hspace{0.75em}2.82^{}$ &             $-2.62^{**}$ &               $-0.36^{}$ &                  $-0.38^{}$ &   $\hspace{0.75em}0.88^{}$ &               $-2.00^{}$ &                $-0.34^{}$ &   $-1.40^{}$ &  $\hspace{0.75em}0.66^{}$ \\
   \midrule
   SST2 &    MRPC &    BERT & $\hspace{0.75em}0.66^{}$ &               $-0.32^{}$ &   $\hspace{0.75em}2.74^{}$ &                  $-0.56^{}$ &   $\hspace{0.75em}1.08^{}$ &   $\hspace{0.75em}0.60^{}$ &                 $-1.62^{}$ &              $-2.10^{*}$ &    $\hspace{0.75em}0.16^{}$ \\
   SST2 &    STSB &    BERT &               $-0.70^{}$ &               $-0.90^{}$ &   $\hspace{0.75em}1.50^{}$ &                  $-0.82^{}$ &   $\hspace{0.75em}1.12^{}$ &                 $-1.56^{}$ &              $-3.66^{***}$ &               $-0.52^{}$ &   $\hspace{0.75em}2.80^{*}$ \\
   SST2 &    QNLI &    BERT & $\hspace{0.75em}0.36^{}$ &               $-0.58^{}$ &   $\hspace{0.75em}0.06^{}$ &    $\hspace{0.75em}0.46^{}$ &   $\hspace{0.75em}0.68^{}$ &                 $-2.30^{}$ &                $-2.80^{*}$ &             $-3.04^{**}$ &                  $-1.28^{}$ \\
   SST2 &     RTE &    BERT & $\hspace{0.75em}1.82^{}$ &              $-2.26^{*}$ &                 $-2.02^{}$ &                  $-0.62^{}$ &   $\hspace{0.75em}0.64^{}$ &                $-2.96^{*}$ &                $-2.00^{*}$ &              $-2.60^{*}$ &                  $-1.30^{}$ \\
   
   \midrule
   MRPC &    STSB &    BERT &               $-2.00^{}$ &               $-0.48^{}$ & $\hspace{0.75em}5.48^{**}$ &                 $-0.94^{}$ &  $\hspace{0.75em}2.22^{*}$ & $\hspace{0.75em}2.42^{*}$ &                $-2.36^{}$ &               $-0.12^{}$ & $\hspace{0.75em}3.42^{**}$ \\
   MRPC &    QNLI &    BERT &               $-0.28^{}$ &               $-1.10^{}$ &  $\hspace{0.75em}4.42^{*}$ &   $\hspace{0.75em}0.50^{}$ &  $\hspace{0.75em}1.76^{*}$ &  $\hspace{0.75em}1.04^{}$ &                $-0.94^{}$ &               $-1.50^{}$ &                 $-0.36^{}$ \\
   MRPC &     RTE &    BERT & $\hspace{0.75em}2.38^{}$ &               $-0.48^{}$ &   $\hspace{0.75em}0.34^{}$ &                 $-0.54^{}$ &  $\hspace{0.75em}1.42^{*}$ &               $-2.38^{*}$ &               $-2.34^{*}$ &              $-2.50^{*}$ &                 $-0.32^{}$ \\

   \midrule
   STSB &    QNLI &    BERT &               $-0.48^{}$ &  $-0.58^{}$ &  $\hspace{0.75em}1.28^{}$ &  $\hspace{0.75em}0.46^{}$ & $\hspace{0.75em}1.44^{*}$ & $\hspace{0.75em}0.14^{}$ &             $-3.18^{**}$ & $\hspace{0.75em}0.96^{}$ & $\hspace{0.75em}1.70^{*}$ \\
   STSB &     RTE &    BERT & $\hspace{0.75em}1.06^{}$ & $-1.88^{*}$ &  $\hspace{0.75em}0.30^{}$ &                $-0.82^{}$ & $\hspace{0.75em}1.70^{*}$ &               $-1.60^{}$ &            $-3.68^{***}$ &               $-0.04^{}$ &  $\hspace{0.75em}2.20^{}$ \\
   
   \midrule
   QNLI &     RTE &    BERT & $\hspace{0.75em}3.46^{}$ & $-1.32^{}$ &      $-2.12^{}$ & $\hspace{0.75em}0.12^{}$ & $\hspace{0.75em}0.36^{}$ & $-3.38^{}$ &             $-2.76^{**}$ & $\hspace{0.75em}0.70^{}$ &               $-1.20^{}$ \\
\bottomrule
\end{tabular}
}
\caption{Interaction effects between GLUE datasets, on BERT. $^{*}$, $^{**}$ and $^{***}$ stand for $p<0.05$, $p<0.01$, and $p<0.001$, respectively, for the ANOVA test of the interaction effect.}
\label{tab:interaction_effects_BERT}
\end{table*}

\section{Discussion}
\paragraph{Checking the effects before using datasets}
Considering that the datasets can show spill-over effects when used independently and interaction effects when used jointly, we call for more careful scrutiny of datasets. While the model developers already have a busy working pipeline, we call for model developers to at least be aware of spill-over effects of the datasets. Considering the possibly negative effects to the models' linguistic abilities, adding datasets to the model's training might not always be beneficial.

\paragraph{Documentation for datasets}
The transparency of model development pipelines can be improved, and better documentation of the data is a crucial improvement area \citep{gebru2021datasheets,paullada_data_2021}. Recently, \citet{pushkarna2022data} described some principles for unified documentation of datasets: flexible, modular, extensible, accessible, and content-agnostic.
The dataset effect can be a module in the dataset documentation. In addition to documenting the basic properties of the datasets, it would be great to also note how the dataset has potential ``spill-over'' effects and ``interaction effects''. This is better done via a joint effort from the AI community.

\paragraph{From data difficulty to ``dataset effects''}
While the difficulty of the dataset is a uni-dimensional score, the effect of datasets can be multi-dimensional. Improving the difficulty of datasets (e.g., by identifying adversarial examples and challenging datasets) has been shown to improve performance \citep{ribeiro_beyond_2020,gardner2020evaluating}. The consideration of multi-dimensional dataset effects can potentially introduce similar benefits.

\section{Conclusion}
We propose a state-vector framework to study \textit{dataset effects}. The framework uses probing classifiers to describe the effects of datasets on the resulting models along multiple linguistic ability dimensions. This framework allows us to identify the individual effects and the interaction effects of a number of datasets. With extensive experiments, we find that the dataset effects are concentrated and characteristic. Additionally, we discuss how the state-vector framework to study dataset effects can improve the dataset curation practice and responsible model development workflow.

\section{Limitations}
\paragraph{Probing tests may not be idealized}
When formulating the framework, we consider idealized probes -- 100\% valid and reliable. In reality, probing tests are unfortunately not ideal yet. We follow the common practice of setting up the probing classifiers to allow fair comparison with their literature. We run the probing experiments on multiple random seeds to reduce the impacts of randomness. 

\paragraph{Model training may not be optimal}
Empirically, the datasets included in our analyses are limited to the fine-tuning stage. Previous work found distinct ``stages'' during the training of DNNs where the DNNs respond to the data samples differently. For example, \citet{shwartz-ziv_opening_2017} referred to the stages as ``drift phase'' and the ``diffusion phase''. The means of the gradients are drastically different between the two stages. \citet{tanzer-etal-2022-memorisation} identified a ``second stage'' where the models do not overfit to noisy data labels.  In the framework of this paper, we consider the \textit{ideal} model training, where our states are defined as the global optimum where the model arrives.

\paragraph{Interaction effects of more than two tasks}
The interaction effect is defined between two tasks. We believe this framework can generalize to more than two tasks, but the empirical verification is left to future work. 

\paragraph{Coverage of experiments}
As the number of datasets we consider increases, the number of experiments in total grows exponentially. It is unrealistic to go through the set of all combinations in our experiments, so we picked some experiments and organized them given the categories of the desired effects (instead of the observed effects) of the datasets. Additional experiments that test the exact interaction effects are left to future works. 
Also, we only considered classification-type tasks in the experiments. While this state-vector framework naturally generalizes to other tasks, including cloze and next-sentence prediction, the empirical observations are left to future works as well. We consider the fine-tuning setting in the experiments. Another setting, language model pre-training, also involves classification-type tasks and usually has larger sets of labels. Our theoretical framework generalizes to the pre-training setting as well.

\bibliography{custom}

\begin{thebibliography}{47}
\expandafter\ifx\csname natexlab\endcsname\relax\def\natexlab#1{#1}\fi

\bibitem[{Aroca-Ouellette and Rudzicz(2020)}]{aroca-ouellette_losses_2020}
Stéphane Aroca-Ouellette and Frank Rudzicz. 2020.
\newblock \href {https://doi.org/10.18653/v1/2020.emnlp-main.403} {On {Losses}
  for {Modern} {Language} {Models}}.
\newblock In \emph{Proceedings of the 2020 {Conference} on {Empirical}
  {Methods} in {Natural} {Language} {Processing} ({EMNLP})}, pages 4970--4981,
  Online. Association for Computational Linguistics.

\bibitem[{Belinkov(2021)}]{belinkov_probing_2021}
Yonatan Belinkov. 2021.
\newblock \href {http://arxiv.org/abs/2102.12452} {Probing {Classifiers}:
  {Promises}, {Shortcomings}, and {Alternatives}}.
\newblock \emph{arXiv:2102.12452}.

\bibitem[{Bentivogli et~al.(2009)Bentivogli, Clark, Dagan, and
  Giampiccolo}]{bentivogli2009fifth}
Luisa Bentivogli, Peter Clark, Ido Dagan, and Danilo Giampiccolo. 2009.
\newblock The fifth pascal recognizing textual entailment challenge.
\newblock \emph{TAC}, 7:8.

\bibitem[{Brown et~al.(2020)Brown, Mann, Ryder, Subbiah, Kaplan, Dhariwal,
  Neelakantan, Shyam, Sastry, Askell, Agarwal, Herbert-Voss, Krueger, Henighan,
  Child, Ramesh, Ziegler, Wu, Winter, Hesse, Chen, Sigler, Litwin, Gray, Chess,
  Clark, Berner, McCandlish, Radford, Sutskever, and
  Amodei}]{brown_language_2020}
Tom Brown, Benjamin Mann, Nick Ryder, Melanie Subbiah, Jared~D Kaplan, Prafulla
  Dhariwal, Arvind Neelakantan, Pranav Shyam, Girish Sastry, Amanda Askell,
  Sandhini Agarwal, Ariel Herbert-Voss, Gretchen Krueger, Tom Henighan, Rewon
  Child, Aditya Ramesh, Daniel Ziegler, Jeffrey Wu, Clemens Winter, Chris
  Hesse, Mark Chen, Eric Sigler, Mateusz Litwin, Scott Gray, Benjamin Chess,
  Jack Clark, Christopher Berner, Sam McCandlish, Alec Radford, Ilya Sutskever,
  and Dario Amodei. 2020.
\newblock \href
  {https://proceedings.neurips.cc/paper/2020/hash/1457c0d6bfcb4967418bfb8ac142f64a-Abstract.html}
  {Language {Models} are {Few}-{Shot} {Learners}}.
\newblock In \emph{Advances in {Neural} {Information} {Processing} {Systems}},
  volume~33, pages 1877--1901.

\bibitem[{Card et~al.(2020)Card, Henderson, Khandelwal, Jia, Mahowald, and
  Jurafsky}]{card_little_2020}
Dallas Card, Peter Henderson, Urvashi Khandelwal, Robin Jia, Kyle Mahowald, and
  Dan Jurafsky. 2020.
\newblock \href {https://aclanthology.org/2020.emnlp-main.745} {With {Little}
  {Power} {Comes} {Great} {Responsibility}}.
\newblock In \emph{Proceedings of the 2020 {Conference} on {Empirical}
  {Methods} in {Natural} {Language} {Processing} ({EMNLP})}, pages 9263--9274,
  Online. Association for Computational Linguistics.

\bibitem[{Cer et~al.(2017)Cer, Diab, Agirre, Lopez-Gazpio, and
  Specia}]{cer-etal-2017-semeval}
Daniel Cer, Mona Diab, Eneko Agirre, I{\~n}igo Lopez-Gazpio, and Lucia Specia.
  2017.
\newblock \href {https://doi.org/10.18653/v1/S17-2001} {{S}em{E}val-2017 task
  1: Semantic textual similarity multilingual and crosslingual focused
  evaluation}.
\newblock In \emph{Proceedings of the 11th International Workshop on Semantic
  Evaluation ({S}em{E}val-2017)}, pages 1--14, Vancouver, Canada. Association
  for Computational Linguistics.

\bibitem[{Conneau and Kiela(2018)}]{conneau-kiela-2018-senteval}
Alexis Conneau and Douwe Kiela. 2018.
\newblock \href {https://aclanthology.org/L18-1269} {{S}ent{E}val: An
  evaluation toolkit for universal sentence representations}.
\newblock In \emph{Proceedings of the Eleventh International Conference on
  Language Resources and Evaluation ({LREC} 2018)}, Miyazaki, Japan. European
  Language Resources Association (ELRA).

\bibitem[{Dagan et~al.(2005)Dagan, Glickman, and Magnini}]{dagan2005pascal}
Ido Dagan, Oren Glickman, and Bernardo Magnini. 2005.
\newblock The pascal recognising textual entailment challenge.
\newblock In \emph{Machine learning challenges workshop}, pages 177--190.
  Springer.

\bibitem[{Devlin et~al.(2019)Devlin, Chang, Lee, and
  Toutanova}]{devlin_BERT_2019}
Jacob Devlin, Ming-Wei Chang, Kenton Lee, and Kristina Toutanova. 2019.
\newblock \href {https://doi.org/10.18653/v1/N19-1423} {{BERT}: {Pre}-training
  of {Deep} {Bidirectional} {Transformers} for {Language} {Understanding}}.
\newblock In \emph{Proceedings of the 2019 {Conference} of the {North}
  {American} {Chapter} of the {Association} for {Computational} {Linguistics}:
  {Human} {Language} {Technologies}, {Volume} 1 ({Long} and {Short} {Papers})},
  pages 4171--4186, Minneapolis, Minnesota. Association for Computational
  Linguistics.

\bibitem[{Dolan and Brockett(2005)}]{dolan-brockett-2005-automatically}
William~B. Dolan and Chris Brockett. 2005.
\newblock \href {https://aclanthology.org/I05-5002} {Automatically constructing
  a corpus of sentential paraphrases}.
\newblock In \emph{Proceedings of the Third International Workshop on
  Paraphrasing ({IWP}2005)}.

\bibitem[{Ethayarajh et~al.(2022)Ethayarajh, Choi, and
  Swayamdipta}]{ethayarajh_2022_dataset}
Kawin Ethayarajh, Yejin Choi, and Swabha Swayamdipta. 2022.
\newblock \href {https://proceedings.mlr.press/v162/ethayarajh22a.html}
  {Understanding dataset difficulty with $\mathcal{V}$-usable information}.
\newblock In \emph{Proceedings of the 39th International Conference on Machine
  Learning}, volume 162 of \emph{Proceedings of Machine Learning Research},
  pages 5988--6008. PMLR.

\bibitem[{Gardner et~al.(2020)Gardner, Artzi, Basmova, Berant, Bogin, Chen,
  Dasigi, Dua, Elazar, Gottumukkala et~al.}]{gardner2020evaluating}
Matt Gardner, Yoav Artzi, Victoria Basmova, Jonathan Berant, Ben Bogin, Sihao
  Chen, Pradeep Dasigi, Dheeru Dua, Yanai Elazar, Ananth Gottumukkala, et~al.
  2020.
\newblock \href {https://arxiv.org/abs/2004.02709} {Evaluating models' local
  decision boundaries via contrast sets}.
\newblock \emph{arXiv preprint arXiv:2004.02709}.

\bibitem[{Gebru et~al.(2021)Gebru, Morgenstern, Vecchione, Vaughan, Wallach,
  Iii, and Crawford}]{gebru2021datasheets}
Timnit Gebru, Jamie Morgenstern, Briana Vecchione, Jennifer~Wortman Vaughan,
  Hanna Wallach, Hal~Daum{\'e} Iii, and Kate Crawford. 2021.
\newblock \href {https://arxiv.org/pdf/1803.09010.pdf} {Datasheets for
  datasets}.
\newblock \emph{Communications of the ACM}, 64(12):86--92.

\bibitem[{Giampiccolo et~al.(2007)Giampiccolo, Magnini, Dagan, and
  Dolan}]{giampiccolo2007third}
Danilo Giampiccolo, Bernardo Magnini, Ido Dagan, and William~B Dolan. 2007.
\newblock The third pascal recognizing textual entailment challenge.
\newblock In \emph{Proceedings of the ACL-PASCAL workshop on textual entailment
  and paraphrasing}, pages 1--9.

\bibitem[{Gupta et~al.(2015)Gupta, Boleda, Baroni, and
  Pad{\'o}}]{gupta-etal-2015-distributional}
Abhijeet Gupta, Gemma Boleda, Marco Baroni, and Sebastian Pad{\'o}. 2015.
\newblock \href {https://doi.org/10.18653/v1/D15-1002} {Distributional vectors
  encode referential attributes}.
\newblock In \emph{Proceedings of the 2015 Conference on Empirical Methods in
  Natural Language Processing}, pages 12--21, Lisbon, Portugal. Association for
  Computational Linguistics.

\bibitem[{Haim et~al.(2006)Haim, Dagan, Dolan, Ferro, Giampiccolo, Magnini, and
  Szpektor}]{haim2006second}
R~Bar Haim, Ido Dagan, Bill Dolan, Lisa Ferro, Danilo Giampiccolo, Bernardo
  Magnini, and Idan Szpektor. 2006.
\newblock The second pascal recognising textual entailment challenge.
\newblock In \emph{Proceedings of the Second PASCAL Challenges Workshop on
  Recognising Textual Entailment}, volume~7, pages 785--794.

\bibitem[{Hewitt and Liang(2019)}]{hewitt-liang-2019-designing}
John Hewitt and Percy Liang. 2019.
\newblock \href {https://doi.org/10.18653/v1/D19-1275} {Designing and
  interpreting probes with control tasks}.
\newblock In \emph{Proceedings of the 2019 Conference on Empirical Methods in
  Natural Language Processing and the 9th International Joint Conference on
  Natural Language Processing (EMNLP-IJCNLP)}, pages 2733--2743, Hong Kong,
  China. Association for Computational Linguistics.

\bibitem[{Hoffmann et~al.(2022)Hoffmann, Borgeaud, Mensch, Buchatskaya, Cai,
  Rutherford, Casas, Hendricks, Welbl, Clark et~al.}]{hoffmann2022training}
Jordan Hoffmann, Sebastian Borgeaud, Arthur Mensch, Elena Buchatskaya, Trevor
  Cai, Eliza Rutherford, Diego de~Las Casas, Lisa~Anne Hendricks, Johannes
  Welbl, Aidan Clark, et~al. 2022.
\newblock \href {https://arxiv.org/pdf/2203.15556.pdf} {Training
  compute-optimal large language models}.
\newblock \emph{arXiv preprint arXiv:2203.15556}.

\bibitem[{Jeon et~al.(2022)Jeon, Aupetit, Shin, Cho, Park, and
  Seo}]{jeon2022sanity}
Hyeon Jeon, Michael Aupetit, DongHwa Shin, Aeri Cho, Seokhyeon Park, and
  Jinwook Seo. 2022.
\newblock \href {https://arxiv.org/abs/2209.10042} {Sanity check for external
  clustering validation benchmarks using internal validation measures}.
\newblock \emph{arXiv preprint arXiv:2209.10042}.

\bibitem[{Levesque et~al.(2012)Levesque, Davis, and
  Morgenstern}]{levesque2012winograd}
Hector Levesque, Ernest Davis, and Leora Morgenstern. 2012.
\newblock The winograd schema challenge.
\newblock In \emph{Thirteenth international conference on the principles of
  knowledge representation and reasoning}.

\bibitem[{Liu et~al.(2021)Liu, Wang, Kasai, Hajishirzi, and
  Smith}]{liu_probing_2021}
Leo~Z. Liu, Yizhong Wang, Jungo Kasai, Hannaneh Hajishirzi, and Noah~A. Smith.
  2021.
\newblock \href {http://arxiv.org/abs/2104.07885} {Probing {Across} {Time}:
  {What} {Does} {RoBERTa} {Know} and {When}?}
\newblock \emph{arXiv:2104.07885 [cs]}.

\bibitem[{Liu et~al.(2019)Liu, Ott, Goyal, Du, Joshi, Chen, Levy, Lewis,
  Zettlemoyer, and Stoyanov}]{liu_RoBERTa_2019}
Yinhan Liu, Myle Ott, Naman Goyal, Jingfei Du, Mandar Joshi, Danqi Chen, Omer
  Levy, Mike Lewis, Luke Zettlemoyer, and Veselin Stoyanov. 2019.
\newblock \href {http://arxiv.org/abs/1907.11692} {{RoBERTa}: {A} {Robustly}
  {Optimized} {BERT} {Pretraining} {Approach}}.
\newblock \emph{arXiv:1907.11692 [cs]}.

\bibitem[{Mosbach et~al.(2020)Mosbach, Khokhlova, Hedderich, and
  Klakow}]{mosbach-etal-2020-interplay-fine}
Marius Mosbach, Anna Khokhlova, Michael~A. Hedderich, and Dietrich Klakow.
  2020.
\newblock \href {https://doi.org/10.18653/v1/2020.findings-emnlp.227} {{O}n the
  {I}nterplay {B}etween {F}ine-tuning and {S}entence-level {P}robing for
  {L}inguistic {K}nowledge in {P}re-trained {T}ransformers}.
\newblock In \emph{Findings of the Association for Computational Linguistics:
  EMNLP 2020}, pages 2502--2516, Online. Association for Computational
  Linguistics.

\bibitem[{Niu et~al.(2022)Niu, Lu, and Penn}]{niu-et-al-2022-BERT}
Jingcheng Niu, Wenjie Lu, and Gerald Penn. 2022.
\newblock \href {https://aclanthology.org/2022.coling-1.278/} {Does {BERT}
  rediscover the classical {NLP} pipeline?}
\newblock In \emph{Proceedings of the 29th International Conference on
  Computational Linguistics}. International Committee on Computational
  Linguistics.

\bibitem[{Paullada et~al.(2021)Paullada, Raji, Bender, Denton, and
  Hanna}]{paullada_data_2021}
Amandalynne Paullada, Inioluwa~Deborah Raji, Emily~M. Bender, Emily Denton, and
  Alex Hanna. 2021.
\newblock \href {https://doi.org/https://doi.org/10.1016/j.patter.2021.100336}
  {Data and its (dis)contents: {A} survey of dataset development and use in
  machine learning research}.
\newblock \emph{Patterns}, 2(11):100336.

\bibitem[{Pimentel and Cotterell(2021)}]{pimentel-cotterell-2021-bayesian}
Tiago Pimentel and Ryan Cotterell. 2021.
\newblock \href {https://doi.org/10.18653/v1/2021.emnlp-main.229} {A {B}ayesian
  framework for information-theoretic probing}.
\newblock In \emph{Proceedings of the 2021 Conference on Empirical Methods in
  Natural Language Processing}, pages 2869--2887, Online and Punta Cana,
  Dominican Republic. Association for Computational Linguistics.

\bibitem[{Pushkarna et~al.(2022)Pushkarna, Zaldivar, and
  Kjartansson}]{pushkarna2022data}
Mahima Pushkarna, Andrew Zaldivar, and Oddur Kjartansson. 2022.
\newblock \href {https://arxiv.org/abs/2204.01075} {{Data Cards: Purposeful and
  Transparent Dataset Documentation for Responsible AI}}.
\newblock \emph{arXiv preprint arXiv:2204.01075}.

\bibitem[{Radford et~al.(2018)Radford, Narasimhan, Salimans, Sutskever, and
  {others}}]{radford_improving_2018}
Alec Radford, Karthik Narasimhan, Tim Salimans, Ilya Sutskever, and {others}.
  2018.
\newblock \href
  {https://s3-us-west-2.amazonaws.com/openai-assets/research-covers/language-unsupervised/language_understanding_paper.pdf}
  {Improving language understanding by generative pre-training}.
\newblock Publisher: OpenAI.

\bibitem[{Rajpurkar et~al.(2016)Rajpurkar, Zhang, Lopyrev, and
  Liang}]{rajpurkar-etal-2016-squad}
Pranav Rajpurkar, Jian Zhang, Konstantin Lopyrev, and Percy Liang. 2016.
\newblock \href {https://doi.org/10.18653/v1/D16-1264} {{SQ}u{AD}: 100,000+
  questions for machine comprehension of text}.
\newblock In \emph{Proceedings of the 2016 Conference on Empirical Methods in
  Natural Language Processing}, pages 2383--2392, Austin, Texas. Association
  for Computational Linguistics.

\bibitem[{Ribeiro et~al.(2020)Ribeiro, Wu, Guestrin, and
  Singh}]{ribeiro_beyond_2020}
Marco~Tulio Ribeiro, Tongshuang Wu, Carlos Guestrin, and Sameer Singh. 2020.
\newblock \href {https://doi.org/10.18653/v1/2020.acl-main.442} {Beyond
  {Accuracy}: {Behavioral} {Testing} of {NLP} {Models} with {CheckList}}.
\newblock In \emph{Proceedings of the 58th {Annual} {Meeting} of the
  {Association} for {Computational} {Linguistics}}, pages 4902--4912, Online.
  Association for Computational Linguistics.

\bibitem[{Rogers et~al.(2020)Rogers, Kovaleva, and
  Rumshisky}]{rogers_primer_2020}
Anna Rogers, Olga Kovaleva, and Anna Rumshisky. 2020.
\newblock \href {https://doi.org/10.1162/tacl_a_00349} {A {Primer} in
  {BERTology}: {What} {We} {Know} {About} {How} {BERT} {Works}}.
\newblock \emph{Transactions of the Association for Computational Linguistics},
  8:842--866.

\bibitem[{Shwartz-Ziv and Tishby(2017)}]{shwartz-ziv_opening_2017}
Ravid Shwartz-Ziv and Naftali Tishby. 2017.
\newblock Opening the black box of deep neural networks via information.
\newblock \emph{arXiv preprint arXiv:1703.00810}.

\bibitem[{Socher et~al.(2013)Socher, Perelygin, Wu, Chuang, Manning, Ng, and
  Potts}]{socher-etal-2013-recursive}
Richard Socher, Alex Perelygin, Jean Wu, Jason Chuang, Christopher~D. Manning,
  Andrew Ng, and Christopher Potts. 2013.
\newblock \href {https://aclanthology.org/D13-1170} {Recursive deep models for
  semantic compositionality over a sentiment treebank}.
\newblock In \emph{Proceedings of the 2013 Conference on Empirical Methods in
  Natural Language Processing}, pages 1631--1642, Seattle, Washington, USA.
  Association for Computational Linguistics.

\bibitem[{Sun et~al.(2021)Sun, Wang, Feng, Ding, Pang, Shang, Liu, Chen, Zhao,
  Lu, and {others}}]{sun_ernie_2021}
Yu~Sun, Shuohuan Wang, Shikun Feng, Siyu Ding, Chao Pang, Junyuan Shang,
  Jiaxiang Liu, Xuyi Chen, Yanbin Zhao, Yuxiang Lu, and {others}. 2021.
\newblock \href {https://arxiv.org/abs/2107.02137} {{ERNIE} 3.0: {Large}-scale
  knowledge enhanced pre-training for language understanding and generation}.
\newblock \emph{arXiv preprint arXiv:2107.02137}.

\bibitem[{Swayamdipta et~al.(2020)Swayamdipta, Schwartz, Lourie, Wang,
  Hajishirzi, Smith, and Choi}]{swayamdipta-etal-2020-dataset}
Swabha Swayamdipta, Roy Schwartz, Nicholas Lourie, Yizhong Wang, Hannaneh
  Hajishirzi, Noah~A. Smith, and Yejin Choi. 2020.
\newblock \href {https://doi.org/10.18653/v1/2020.emnlp-main.746} {Dataset
  cartography: Mapping and diagnosing datasets with training dynamics}.
\newblock In \emph{Proceedings of the 2020 Conference on Empirical Methods in
  Natural Language Processing (EMNLP)}, pages 9275--9293, Online. Association
  for Computational Linguistics.

\bibitem[{T{\"a}nzer et~al.(2022)T{\"a}nzer, Ruder, and
  Rei}]{tanzer-etal-2022-memorisation}
Michael T{\"a}nzer, Sebastian Ruder, and Marek Rei. 2022.
\newblock \href {https://doi.org/10.18653/v1/2022.acl-long.521} {Memorisation
  versus generalisation in pre-trained language models}.
\newblock In \emph{Proceedings of the 60th Annual Meeting of the Association
  for Computational Linguistics (Volume 1: Long Papers)}, pages 7564--7578,
  Dublin, Ireland. Association for Computational Linguistics.

\bibitem[{Tenney et~al.(2019)Tenney, Das, and Pavlick}]{tenney_BERT_2019}
Ian Tenney, Dipanjan Das, and Ellie Pavlick. 2019.
\newblock \href {https://doi.org/10.18653/v1/P19-1452} {{BERT} {Rediscovers}
  the {Classical} {NLP} {Pipeline}}.
\newblock In \emph{Proceedings of the 57th {Annual} {Meeting} of the
  {Association} for {Computational} {Linguistics}}, pages 4593--4601, Florence,
  Italy. Association for Computational Linguistics.

\bibitem[{Villalobos et~al.(2022)Villalobos, Sevilla, Heim, Besiroglu,
  Hobbhahn, and Ho}]{villalobos_will_2022}
Pablo Villalobos, Jaime Sevilla, Lennart Heim, Tamay Besiroglu, Marius
  Hobbhahn, and Anson Ho. 2022.
\newblock \href {http://arxiv.org/abs/2211.04325} {Will we run out of data?
  {An} analysis of the limits of scaling datasets in {Machine} {Learning}}.

\bibitem[{Voita and Titov(2020)}]{voita-titov-2020-information}
Elena Voita and Ivan Titov. 2020.
\newblock \href {https://doi.org/10.18653/v1/2020.emnlp-main.14}
  {Information-theoretic probing with minimum description length}.
\newblock In \emph{Proceedings of the 2020 Conference on Empirical Methods in
  Natural Language Processing (EMNLP)}, pages 183--196, Online. Association for
  Computational Linguistics.

\bibitem[{Wang et~al.(2018)Wang, Singh, Michael, Hill, Levy, and
  Bowman}]{wang-etal-2018-glue}
Alex Wang, Amanpreet Singh, Julian Michael, Felix Hill, Omer Levy, and Samuel
  Bowman. 2018.
\newblock \href {https://doi.org/10.18653/v1/W18-5446} {{GLUE}: A multi-task
  benchmark and analysis platform for natural language understanding}.
\newblock In \emph{Proceedings of the 2018 {EMNLP} Workshop {B}lackbox{NLP}:
  Analyzing and Interpreting Neural Networks for {NLP}}, pages 353--355,
  Brussels, Belgium. Association for Computational Linguistics.

\bibitem[{Warstadt et~al.(2018)Warstadt, Singh, and
  Bowman}]{warstadt2018neural}
Alex Warstadt, Amanpreet Singh, and Samuel~R Bowman. 2018.
\newblock Neural network acceptability judgments.
\newblock \emph{arXiv preprint arXiv:1805.12471}.

\bibitem[{Wei et~al.(2021)Wei, Bosma, Zhao, Guu, Yu, Lester, Du, Dai, and
  Le}]{wei2021finetuned}
Jason Wei, Maarten Bosma, Vincent~Y Zhao, Kelvin Guu, Adams~Wei Yu, Brian
  Lester, Nan Du, Andrew~M Dai, and Quoc~V Le. 2021.
\newblock \href {https://arxiv.org/abs/2109.01652} {Finetuned language models
  are zero-shot learners}.
\newblock \emph{arXiv preprint arXiv:2109.01652}.

\bibitem[{Weller et~al.(2022)Weller, Seppi, and Gardner}]{weller-etal-2022-use}
Orion Weller, Kevin Seppi, and Matt Gardner. 2022.
\newblock \href {https://doi.org/10.18653/v1/2022.acl-short.30} {When to use
  multi-task learning vs intermediate fine-tuning for pre-trained encoder
  transfer learning}.
\newblock In \emph{Proceedings of the 60th Annual Meeting of the Association
  for Computational Linguistics (Volume 2: Short Papers)}, pages 272--282,
  Dublin, Ireland. Association for Computational Linguistics.

\bibitem[{Wolf et~al.(2020)Wolf, Debut, Sanh, Chaumond, Delangue, Moi, Cistac,
  Rault, Louf, Funtowicz, Davison, Shleifer, von Platen, Ma, Jernite, Plu, Xu,
  Le~Scao, Gugger, Drame, Lhoest, and Rush}]{wolf-etal-2020-transformers}
Thomas Wolf, Lysandre Debut, Victor Sanh, Julien Chaumond, Clement Delangue,
  Anthony Moi, Pierric Cistac, Tim Rault, Remi Louf, Morgan Funtowicz, Joe
  Davison, Sam Shleifer, Patrick von Platen, Clara Ma, Yacine Jernite, Julien
  Plu, Canwen Xu, Teven Le~Scao, Sylvain Gugger, Mariama Drame, Quentin Lhoest,
  and Alexander Rush. 2020.
\newblock \href {https://doi.org/10.18653/v1/2020.emnlp-demos.6} {Transformers:
  State-of-the-art natural language processing}.
\newblock In \emph{Proceedings of the 2020 Conference on Empirical Methods in
  Natural Language Processing: System Demonstrations}, pages 38--45, Online.
  Association for Computational Linguistics.

\bibitem[{Xu et~al.(2020)Xu, Zhao, Song, Stewart, and Ermon}]{xu_theory_2020}
Yilun Xu, Shengjia Zhao, Jiaming Song, Russell Stewart, and Stefano Ermon.
  2020.
\newblock \href {https://openreview.net/pdf?id=r1eBeyHFDH} {A theory of usable
  information under computational constraints}.
\newblock \emph{ICLR}.

\bibitem[{Zhu et~al.(2022{\natexlab{a}})Zhu, Shahtalebi, and
  Rudzicz}]{zhu_predicting_2022}
Zining Zhu, Soroosh Shahtalebi, and Frank Rudzicz. 2022{\natexlab{a}}.
\newblock \href {https://arxiv.org/abs/2210.07352} {Predicting fine-tuning
  performance with probing}.
\newblock In \emph{{EMNLP}}.

\bibitem[{Zhu et~al.(2022{\natexlab{b}})Zhu, Wang, Li, and
  Rudzicz}]{zhu_data_2022}
Zining Zhu, Jixuan Wang, Bai Li, and Frank Rudzicz. 2022{\natexlab{b}}.
\newblock \href {https://aclanthology.org/2022.findings-acl.326/} {On the data
  requirements of probing}.
\newblock In \emph{Findings of the {Association} of {Computational}
  {Linguistics}}. Association for Computational Linguistics.

\end{thebibliography}
\bibliographystyle{acl_natbib}

\newpage
.
\newpage

\appendix

\section{Appendix}
\subsection{Figure illustrating the experimental setup}
\begin{figure}[h]
    \centering
    \includegraphics[width=\linewidth]{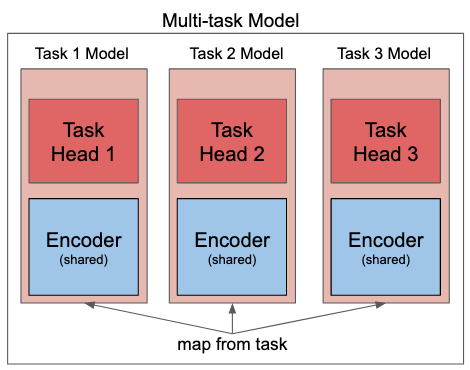}
    \caption{Multitask model architecture for fine-tuning experiments.\footnotemark}
    \label{fig:finetune_setup}
\end{figure}
\footnotetext{Experimental setup based on this  \href{https://colab.research.google.com/github/zphang/zphang.github.io/blob/master/files/notebooks/Multi_task_Training_with_Transformers_NLP.ipynb}{guide} }

\subsection{Additional details on multitask settings}
\label{subsec:appendix-multitask-settings}

Consider the number of different multitask fine-tuning settings possible assuming a constant random seed. 
If we fine-tune encoders task-by-task (e.g., first COLA, then MRPC) such that order matters, then this problem is equivalent to the number of ordered subsets that can be formed from the set $T=\{$COLA, SST2, MRPC, STSB, QNLI, RTE$\}$. This computes to $\sum_{n=0}^{6}\frac{6!}{(6-n)!} = 1957$ models per encoder, or $1957\cdot2=3914$ total models. Note the lower bound of the summation is 0, as the empty set corresponds to the baseline model (i.e., no fine-tuning). 

If we discount ordering effects, then the problem reduces to the number of subsets $T$ contains. This evaluates to $2^{6}=64$ models per encoder or $64\cdot2=128$ total models, which results in far fewer experiments. Ordering effects can be discounted by aggregating training samples of each task, then randomly sampling from this combined dataset when training. Each sample will have a tag indicating which task it pertains to, so it can be redirected to the correct classification head during training.    

Note that we initially assume a constant random seed. Later, we expanded to five random seeds (42, 1, 1234, 123, 10) to allow statistical significance testing. This increases total models to $128\cdot5=640$, which requires excessive compute resources. There is also the need to organize experiments better to illustrate potential individual and interaction effects clearly. 

To address these issues, we impose the following condition: \textbf{experiments must constitute of equal task counts per task group OR all tasks must belong to the same task group.} Recall the task groups from Section \ref{section:finetune} to be single-sentence, similarity and paraphrase, and inference. This enables us to organize the experiments by marking the states as follows: 
\begin{itemize}[nosep]
    \item $I$: The initial state.
    \item $A$: The model is trained on one dataset.
    \item $B$: The model is trained on two datasets from the same group.
    \item $C$: The model is trained on two datasets from different groups.
    \item $D$: The model is trained on three datasets from different groups.
    \item $E$: The model is trained on four tasks from two groups (two per group).
    \item $F$: The model is trained on six tasks from three groups (two per group)
\end{itemize}

As demonstrated in Table \ref{tab:exp-states-table}, the total number of models we need to train is reduced to $34\cdot2\cdot5=340$.  
\begin{table}[t]
    \centering
    \resizebox{0.8\linewidth}{!}{
    \begin{tabular}{cccc}
    \toprule 
        Marker & N. Groups & N. Tasks & N. Experiments \\ \midrule 
        $I$ & 0 & 0 & 1 \\
        $A$ & 1 & 1 & 6 \\
        $B$ & 1 & 2 & 3 \\
        $C$ & 2 & 1 & 12 \\
        $D$ & 3 & 1 & 8 \\
        $E$ & 2 & 2 & 3 \\
        $F$ & 3 & 2 & 1\\ \bottomrule
    \end{tabular}}
    \caption{Multitask states with counts of groups, tasks per group, and experiments per encoder.}
    \label{tab:exp-states-table}
\end{table}
Designing experiments this way allows framing the dataset effects. 

The individual effects can be framed as transitions between marked states (i.e. adding some task to one state yields another state). For example, $I \rightarrow A$ can reflect the individual effect of a dataset, conditioned on the ``no-fine-tuning initial state'' $I$. $A\rightarrow B$ denotes the individual effect of a dataset $X$, conditioned on the initial state that contains a dataset ($Y$), where $Y$ is in the same group as $X$.

The interaction effects can be framed as combinations between multiple states. For example, two states labeled $A$ (with dataset $X$ and $Y$, respectively) and a state labeled $B$ (with datasets $[X,Y]$) can jointly define the interaction between the datasets $X$ and $Y$. This can be written as $B=A+A$.

This labeling mechanism of the states can support the following effects:  
\begin{itemize}[nosep]
   \item Individual effects: $I \rightarrow A$, $A \rightarrow B$, $A \rightarrow C$, $C \rightarrow D$
   \item Interaction effects: $B=A+A$, $C=A+A$, $D=A+C$, $E=B+B$, $E=C+C$, $F=B+E$, $F=D+D$  
\end{itemize}  

Note that although it is possible to compute interaction effects of more than two datasets, we chose not to focus on these cases as it adds an extra layer of complexity. Hence, we only consider the following interaction effects: $B=A+A$, $C=A+A$. This means we don't need to train models for states $E$ and $F$, reducing total experiments to $30\cdot2\cdot5=300$.

\subsection{Additional math motivation}
\label{subsec:abelian-group-motivation}
Here we provide some additional mathematical motivations for the proposed state-vector framework: dataset effects form an Abelian group.

Given a reference state $S_{I}$, the collection of all possible dataset effects $\mathbf{E}(X) \in \mathcal{E}$ forms an additive Abelian group. Here we show that $\mathcal{E}$ satisfies the requirements.

\textit{Existence of zero}. We already know that the identity element is $\mathbf{0}\in \mathbb{R}^K$. Intuitively, the identity element corresponds to ``no effect'' for this dataset.

\textit{Existence and closure of addition}. The addition operation refers to the vector addition. Since $\mathcal{E}$ is defined under $\mathbb{R}^{K}$, it is closed under addition. Note that due to the interaction effect, addition does \textit{not} refer to applying two datasets together to the ``bucket'' of data for multitask training.

\textit{Existence and closure of negation}. The negation operation refers to flipping the direction of a vector in $\mathcal{E}$. Empirically, negating $\mathbf{E}(X)$ involves a counterfactual query of the effect of a dataset: if $X$ were not applied, what would have been the effect on the state of the model? 

\textit{Associativity and commutativity}. Vector addition satisfies associativity: $(\mathbf{E}(X)+\mathbf{E}(Y))+\mathbf{E}(Z) = \mathbf{E}(X)+(\mathbf{E}(Y))+\mathbf{E}(Z))$ and commutativity: $\mathbf{E}(X) + \mathbf{E}(Y) = \mathbf{E}(Y) + \mathbf{E}(X)$. 
$\qed$

\subsection{On the equivalence between formulations of interaction effects}
\label{subsec:interaction-effect-formulations-equivalence}
Plugging in the indicator variables into Eq. \ref{eq:linreg-interaction} yields the following equation (writing in matrix form):
\begin{align}
\begin{bmatrix}
    \mathbf{S}(I) \\ \mathbf{S}([X,I]) \\ \mathbf{S}([Y,I]) \\ \mathbf{S}([X,Y,I])
\end{bmatrix} = 
\begin{bmatrix}
    1 & 0 & 0 & 0 \\
    1 & 1 & 0 & 0 \\
    1 & 0 & 1 & 0 \\
    1 & 1 & 1 & 1
\end{bmatrix} 
\begin{bmatrix}
    \beta_0 \\ \beta_1 \\ \beta_2 \\ \beta_3
\end{bmatrix} + \epsilon.
\end{align}
The standard variable elimination operations give us an expression for the expression for the interaction effect parameter $\beta_3$:
\begin{align}
    \beta_3 = \mathbf{S}([X,Y,I]) - \mathbf{S}([X,I]) - \mathbf{S}([Y,I]) + \mathbf{S}(I),
\end{align}
which exactly recovers the definition for $\textrm{Int}(X,Y)$ using the equivalent formulation (Eq. \ref{eq:interaction-effect-equiv}).

\subsection{Additional experiment results}
Tables \ref{tab:ind-effects-sst2} -- \ref{tab:interaction_effects_BERT} present some additional experiment results.
\begin{table*}
    \centering 
    \resizebox{\linewidth}{!}{
\begin{tabular}{llllllllllll}
\toprule
Dataset &  Reference &    Model &         Length &                     Depth &       TopConst &    BigramShift &                     Tense &               SubjNumber &      ObjNumber &                    OddManOut &                  CoordInv \\
\midrule
SST2    &     $I$     &  BERT &  $-4.58^{***}$ &   $\hspace{0.75em}0.3^{}$ &    $-4.94^{*}$ &  $-1.86^{***}$ &               $-0.88^{*}$ &              $-0.86^{*}$ &       $\hspace{0.75em}0.0^{}$ &  $\hspace{0.75em}1.22^{***}$ &                  $\hspace{0.75em}0.0^{}$ \\
SST2    &       COLA &  BERT &    $-4.42^{*}$ &                 $-1.7^{}$ &      $-2.0^{}$ &    $-0.78^{*}$ &                $-0.44^{}$ &               $-0.62^{}$ &     $-0.36^{}$ &                   $-0.12^{}$ &  $\hspace{0.75em}1.52^{}$ \\
SST2    &       MRPC &  BERT &    $-3.92^{*}$ &                $-0.02^{}$ &      $-2.2^{}$ &    $-2.42^{*}$ &   $\hspace{0.75em}0.2^{}$ &               $-0.26^{}$ &     $-1.62^{}$ &                   $-0.88^{}$ &  $\hspace{0.75em}0.16^{}$ \\
SST2    &       STSB &  BERT &    $-5.28^{*}$ &                 $-0.6^{}$ &     $-3.44^{}$ &  $-2.68^{***}$ &  $\hspace{0.75em}0.24^{}$ &              $-2.42^{*}$ &  $-3.66^{***}$ &      $\hspace{0.75em}0.7^{}$ &  $\hspace{0.75em}2.8^{*}$ \\
SST2    &       QNLI &  BERT &    $-4.22^{*}$ &                $-0.28^{}$ &  $-4.88^{***}$ &      $-1.4^{}$ &                 $-0.2^{}$ &              $-3.16^{*}$ &     $-2.8^{*}$ &                   $-1.82^{}$ &                $-1.28^{}$ \\
SST2    &        RTE &  BERT &    $-2.76^{*}$ &                $-1.96^{}$ &  $-6.96^{***}$ &  $-2.48^{***}$ &                $-0.24^{}$ &              $-3.82^{*}$ &     $-2.0^{*}$ &                   $-1.38^{}$ &                 $-1.3^{}$ \\
SST2    &  MRPC QNLI &  BERT &     $-2.52^{}$ &  $\hspace{0.75em}0.34^{}$ &    $-4.92^{*}$ &    $-2.06^{*}$ &                $-0.66^{}$ &                $-1.9^{}$ &     $-1.46^{}$ &                    $-1.5^{}$ &                $-0.58^{}$ \\
SST2    &   MRPC RTE &  BERT &    $-4.52^{*}$ &                $-0.84^{}$ &      $-2.5^{}$ &    $-2.18^{*}$ &                $-0.58^{}$ &  $\hspace{0.75em}0.5^{}$ &     $-1.58^{}$ &                  $-1.38^{*}$ &                $-0.36^{}$ \\
SST2    &  STSB QNLI &  BERT &    $-3.06^{*}$ &  $\hspace{0.75em}0.74^{}$ &     $-2.66^{}$ &    $-1.78^{*}$ &  $\hspace{0.75em}0.36^{}$ &               $-2.14^{}$ &     $-1.36^{}$ &                    $-0.4^{}$ &   $\hspace{0.75em}0.8^{}$ \\
SST2    &   STSB RTE &  BERT &    $-5.48^{*}$ &                $-1.04^{}$ &    $-4.66^{*}$ &    $-2.24^{*}$ &                $-0.22^{}$ &              $-2.86^{*}$ &    $-2.48^{*}$ &                   $-2.36^{}$ &                $-1.12^{}$ \\
\midrule
SST2    &     $I$     &  RoBERTa &  $-5.44^{***}$ &    $-2.84^{*}$ &              $-5.7^{***}$ &  $-8.76^{***}$ &              $-4.0^{***}$ &  $-7.28^{***}$ &  $-8.06^{***}$ &    $-2.32^{*}$ &             $-3.76^{***}$ \\
SST2    &       COLA &  RoBERTa &  $-8.28^{***}$ &     $-1.9^{*}$ &             $-8.14^{***}$ &    $-1.46^{*}$ &               $-1.84^{*}$ &    $-4.44^{*}$ &    $-3.86^{*}$ &  $-2.56^{***}$ &                $-1.94^{}$ \\
SST2    &       MRPC &  RoBERTa &  $-9.18^{***}$ &     $-1.88^{}$ &             $-8.42^{***}$ &     $-3.12^{}$ &                $-2.02^{}$ &     $-2.88^{}$ &  $-4.32^{***}$ &     $-1.44^{}$ &             $-4.26^{***}$ \\
SST2    &       STSB &  RoBERTa &   $-8.3^{***}$ &    $-2.78^{*}$ &               $-5.34^{*}$ &     $-2.12^{}$ &   $\hspace{0.75em}0.4^{}$ &     $-3.08^{}$ &  $-4.46^{***}$ &     $-0.88^{}$ &               $-2.32^{*}$ \\
SST2    &       QNLI &  RoBERTa &    $-5.26^{*}$ &     $-1.82^{}$ &                $-5.52^{}$ &  $-6.64^{***}$ &                $-1.36^{}$ &      $-4.2^{}$ &    $-4.22^{*}$ &     $-1.72^{}$ &                $-0.94^{}$ \\
SST2    &        RTE &  RoBERTa &  $-5.64^{***}$ &     $-1.36^{}$ &  $\hspace{0.75em}2.44^{}$ &  $-2.98^{***}$ &  $\hspace{0.75em}0.68^{}$ &     $-1.72^{}$ &     $-2.58^{}$ &     $-1.82^{}$ &  $\hspace{0.75em}2.2^{*}$ \\
SST2    &  MRPC QNLI &  RoBERTa &     $-7.1^{*}$ &    $-2.26^{*}$ &             $-7.94^{***}$ &     $-4.6^{*}$ &               $-2.42^{*}$ &      $-1.8^{}$ &     $-2.38^{}$ &     $-3.2^{*}$ &                $-2.28^{}$ \\
SST2    &   MRPC RTE &  RoBERTa &    $-4.44^{*}$ &     $-0.58^{}$ &               $-6.96^{*}$ &     $-4.6^{*}$ &                $-1.22^{}$ &    $-6.96^{*}$ &    $-4.46^{*}$ &     $-1.46^{}$ &                 $-3.9^{}$ \\
SST2    &  STSB QNLI &  RoBERTa &    $-7.84^{*}$ &     $-0.86^{}$ &                 $-3.5^{}$ &    $-5.92^{*}$ &                 $-1.7^{}$ &     $-2.36^{}$ &     $-2.42^{}$ &     $-1.96^{}$ &                $-1.94^{}$ \\
SST2    &   STSB RTE &  RoBERTa &  $-9.76^{***}$ &  $-2.34^{***}$ &               $-9.72^{*}$ &    $-4.14^{*}$ &                $-2.18^{}$ &      $-4.5^{}$ &     $-2.88^{}$ &     $-1.18^{}$ &                $-1.64^{}$ \\
\bottomrule
\end{tabular}
    }
    \caption{Individual effects of SST2 dataset with different reference states. $^{*}$, $^{**}$ and $^{***}$ stand for $p<0.05$, $p<0.01$, and $p<0.001$, respectively, for two-sample $t$-test with $dof=8$.}
    \label{tab:ind-effects-sst2}
\end{table*}

\end{document}